\documentclass[article, nojss]{jss}\usepackage[]{graphicx}\usepackage[]{xcolor}
\makeatletter
\def\maxwidth{ %
  \ifdim\Gin@nat@width>\linewidth
    \linewidth
  \else
    \Gin@nat@width
  \fi
}
\makeatother

\definecolor{fgcolor}{rgb}{0.345, 0.345, 0.345}

\usepackage{framed}
\makeatletter
 {\par\unskip\endMakeFramed%
 \at@end@of@kframe}
\makeatother

\definecolor{shadecolor}{rgb}{.97, .97, .97}
\definecolor{messagecolor}{rgb}{0, 0, 0}
\definecolor{warningcolor}{rgb}{1, 0, 1}
\definecolor{errorcolor}{rgb}{1, 0, 0}
\usepackage{alltt}

\usepackage{thumbpdf,lmodern}

\usepackage{framed}
\usepackage{amsmath,amsthm}
\usepackage{amsfonts}
\usepackage{amssymb}
\usepackage{rotating}
\usepackage{orcidlink}
\usepackage{enumerate}
\usepackage{float} % to allow H for figure position
\usepackage{longtable}
\usepackage[tableposition=below]{caption}
\usepackage{subcaption}

\usepackage{algorithm}
\usepackage{algorithmic}

\usepackage{lmodern}

\theoremstyle{definition}
\newtheorem{definition}{Definition}

\author{Susanne Dandl~\orcidlink{0000-0003-4324-4163}\\%Department of Statistics \\
LMU Munich  \\
MCML\\
   \And Andreas Hofheinz\\ %Department of Statistics \\
   LMU Munich \\
   \And Martin Binder \\% Department of Statistics \\
   LMU Munich \\
   MCML\\
   \AND Bernd Bischl~\orcidlink{0000-0001-6002-6980} \\%Department of Statistics \\
   LMU Munich \\
   MCML\\
   \And Giuseppe Casalicchio~\orcidlink{0000-0001-5324-5966} \\%Department of Statistics \\
   LMU Munich \\
   MCML}
\Plainauthor{Susanne Dandl, Andreas Hofheinz, Martin Binder, Bernd Bischl, Giuseppe Casalicchio}

\title{\pkg{counterfactuals}: An \proglang{R} Package for Counterfactual Explanation Methods}
\Plaintitle{counterfactuals: An R package for counterfactual explanation methods}
\Shorttitle{\pkg{counterfactuals}: An \proglang{R} Package for Counterfactual Explanations}

\Abstract{
    Counterfactual explanation methods provide information on how feature values of individual observations must be changed to obtain a desired prediction. %  so that a black box model predicts a desired value.
    Despite the increasing amount of proposed methods in research, only a few implementations exist whose interfaces and requirements vary widely. 
    In this work, we introduce the \pkg{counterfactuals} \proglang{R} package, which provides a modular and unified \pkg{R6}-based interface for counterfactual explanation methods. 
    We implemented three existing counterfactual explanation methods and propose some optional methodological extensions to generalize these methods to different scenarios and to make them more comparable.
    We explain the structure and workflow of the package using real use cases and show how to integrate additional counterfactual explanation methods into the package.
    In addition, we compared the implemented methods for a variety of models and datasets with regard to the quality of their counterfactual explanations and their runtime behavior.
}

\Plainkeywords{counterfactual explanations, interpretable machine learning, R}

\Address{
  Susanne Dandl \\
  Insitut f\"ur Statistik \\ Ludwig-Maximilians-Universit\"at M\"unchen, Germany \\
  Ludwigstr. 33, 80539 Munich, Germany \\
  Munich Center for Machine Learning (MCML), Germany \\
  E-mail: \email{Susanne.Dandl@stat.uni-muenchen.de}
}
\IfFileExists{upquote.sty}{\usepackage{upquote}}{}
\begin{document}

%% -- Introduction -------------------------------------------------------------

%% - In principle "as usual".
%% - But should typically have some discussion of both _software_ and _methods_.
%% - Use \proglang{}, \pkg{}, and \code{} markup throughout the manuscript.
%% - If such markup is in (sub)section titles, a plain text version has to be
%%   added as well.
%% - All software mentioned should be properly \cite-d.
%% - All abbreviations should be introduced.
%% - Unless the expansions of abbreviations are proper names (like "Journal
%%   of Statistical Software" above) they should be in sentence case (like
%%   "generalized linear models" below).

\section[Introduction]{Introduction and related work} \label{sec:intro}

In recent years, counterfactual explanation methods have emerged as valuable techniques 
for explaining single predictions of black-box models.
Denied loan applications serve as a common example; here, a counterfactual explanation 
(or \textit{counterfactual} for short) could be: ``You were denied a loan because your annual 
income was \pounds30,000. If your income had been \pounds45,000, you would have been offered a loan'' \citep{ref-wachter2017counterfactual}.
More generally, counterfactuals address questions of the form:
``For input $\mathbf{x^{\star}}$, the model predicted $y$. What needs to be changed 
in $\mathbf{x^{\star}}$ so that the model predicts a desired outcome $y'$
instead?''$.$

One advantage of counterfactuals is their human-friendly interpretability: as they simply suggest feature changes to obtain a desired outcome, they are comprehensible even to non-experts \citep{ref-molnar2019}. 
%In addition, counterfactuals give reasons why a model made a certain decision and provide a potential basis to challenge unfair outcomes \citep{ref-wachter2017counterfactual}. 
In addition, counterfactual scenarios can help to detect biases of individual predictions \citep{ref-wachter2017counterfactual}. 
There are several ways to change features to obtain a desired outcome, 
but not all of them are feasible. Therefore,
counterfactual methods that provide multiple (reasonable) counterfactuals and allow the user
to assess their usefulness using domain knowledge are preferable \citep{ref-dandl2020}.
Counterfactual explanations are related to adversarial examples \citep{ref-szegedy2014ae}, but the latter aim to deceive a model instead of explaining it \citep{ref-freiesleben2021ae}.

Over the past few years, a variety of counterfactual explanation methods have been proposed.
Overviews are given in \cite{ref-verma2020counterfactual}, \cite{ref-karimi2021algorithmic}, and \cite{ref-stepin2021review}. 
Most of the methods focus on classification models and use either optimization techniques or heuristic rules to search for counterfactuals.
Existing methods are either model-specific in the sense that they are only applicable to certain model classes (e.g., linear or tree-based models) or model-agnostic, i.e., they are applicable to arbitrary models.
%The methods could be either model-specific or model-agnostic. The former only handle specific models, for example, differential, linear or tree-based models. The latter do not put any restrictions on the underlying model. 
Furthermore, the methods differ in whether and to what extent access to the underlying data is necessary, the number of counterfactuals they return, and the properties of counterfactuals targeted by a method (e.g., sparsity or actionability). 
We will present the most frequently targeted properties in Definition~\ref{def:cfe}.
Counterfactual explanation methods which explicitly target actionable feature changes are also called \textit{recourse} \citep{ref-verma2020counterfactual}.

Despite the increasing amount of proposed counterfactual methods in research, the current software landscape is rather sparse.
To the best of our knowledge, the only counterfactual methods available in \proglang{R} \citep{ref-RCore2021} as dedicated packages are \textit{MOC} \citep{ref-dandl2020, ref-dandl2020github} 
and Feature Tweaking \citep{ref-tolomei2017interpretable, ref-kato2018github}. Feature Tweaking is a model-specific method tailored to random forests and its \proglang{R} implementation only allows forests specifically trained with the \pkg{randomForest} package. In contrast, \textit{MOC} is a model-agnostic method and its implementation allows all regression or classification models fitted with popular toolboxes such as \pkg{caret} \citep{ref-kuhn2021caret} and \pkg{mlr3} \citep{ref-mlr32019}. Models of other packages can also be processed using a wrapper function. %whose implementation is at an early stage (details are given in Section~\ref{extension-of-the-package}).
%In \proglang{Python} \citep{ref-van1995python}, libraries for \textit{NICE} \citep{ref-brughmans2021nice, ref-brughmans2021github} and
%DiCE \citep{ref-mothilal2020explaining, ref-mothilal2021github} exist which both allow models fitted with sklearn (and PyTorch for DiCE).
%but has been shown to be inferior to \textit{MOC}  \citep{ref-dandl2020}.
%Prototypes of MACE \citep{ref-karimi2020model} and MINT \citep{ref-karimi2021algorithmic} are available in one repository \citep{ref-karimi2021github} but limited to certain datasets
%and models (logistic regression, classification tree, random forest and neural network fitted with \pkg{scikit-learn} \citep{ref-pedregosa2011scikit-learn}.
%Implementations of LORE
%\citep{ref-guidotti2018local, ref-guidotti2018github} 
%and Feature
%Tweaking \citep{ref-tolomei2017interpretable, ref-ishihara2018github} 
%do exist but not yet as self-contained and installable libraries.
%Recourse of
%\cite{ref-ustun2019actionable} \citep{ref-ustun2020github}
%and an extended version of the method of
%\cite{ref-wachter2017counterfactual}
%are implemented in the \pkg{alibi} library \citep{ref-klaise2021github}.
%The former is limited to linear models and the latter requires
%differentiable models. 
In \proglang{Python} \citep{ref-van1995python}, the \pkg{CARLA} library \citep{ref-pawelczyk2021carla} provides a variety of (model-agnostic and model-specific)
counterfactual explanation methods for classification models.
\pkg{CARLA} currently calls the original \proglang{Python} implementations of the methods, which often only allow models of specific ML libraries as an input.
Furthermore, a library for the model-agnostic method \textit{NICE} \citep{ref-brughmans2021nice, ref-brughmans2021github} exists which could process all models fitted with \pkg{scikit-learn} \citep{ref-pedregosa2011scikit-learn}. 
Implementations of the methods MACE \citep{ref-karimi2020model}, MINT \citep{ref-karimi2021algorithmic} and LORE \citep{ref-guidotti2018local} are available \citep{ref-karimi2021github, ref-guidotti2018github}, but these are only meant to reproduce the experiments of the original paper, and are therefore limited to certain datasets and models.
Apart from MOC, all the mentioned methods are not capable of returning multiple counterfactuals (in one run).

In summary, existing implementations are predominantly available in \proglang{Python} in different repositories or libraries and at different stages of development.
\proglang{R} users can only access a limited number of methods, and the usability and comparability of these methods are severely limited because there is no common user interface.
% Because the available packages do not have a common user interface, it is difficult to compare different methods. 
Most \proglang{Python} libraries only allow methods for classification models and focus primarily on methods returning a single counterfactual.

%In summary, existing implementations are mostly at an early stage, predominantly available in \proglang{Python}, often tailored to specific problems, and usually limited to returning single counterfactuals. Finally, because the available packages do not have a common user interface, it is difficult to compare different methods. 
%Different interfaces and specific requirements of the available implementations make pose an additional challenge; 
%for example, while the implementation of DiCE requires the model to be encapsulated in a model wrapper class, the implementation of \textit{NICE} requires the prediction function to be extracted from the model.

\textbf{Contributions}: With the \pkg{counterfactuals} package, we offer the first \proglang{R} package that provides a user-friendly and unified interface for model-specific as well as model-agnostic counterfactual explanation methods.
Therefore, it complements other \proglang{R}-based toolkits for interpreting machine learning models such as \pkg{IML} \citep{ref-molnar2019} and \pkg{DALEX} \citep{ref-DALEX2018}.
The package provides common functionalities to evaluate and visualize counterfactuals of diverse methods. 
It is flexible enough to be easily extended by other counterfactual methods for classification or regression models. 
Currently, the package provides three counterfactual explanations methods.
We discuss some (optional) extensions we have made to these methods: first, to generalize them to diverse scenarios (for example, to regression models or multiclass classifiers), and second, to improve their comparability, for example, by letting the two methods, that return only one counterfactual, return several ones just like the third method.
Our work is therefore one of the few that explicitly advocates methods that simultaneously generate multiple, qualitatively comparable counterfactuals rather than a single one.
We are also among the first to provide an evaluation approach for \textit{different sized} sets of counterfactuals by comparing the three implemented methods in a benchmark study.
In contrast, previous work primarily focused on one counterfactual per method \citep{ref-deoliveira21benchmark, ref-pawelczyk2021carla, ref-moreiraetal2022benchmark}.
%This should help users to choose an appropriate method for their use case. 
Because the package and benchmark study code are freely available, we encourage readers to add counterfactual approaches to our \proglang{R} package and compare them to the ones that have already been implemented. 

In the upcoming section, we present the three currently implemented methods.
In Section~\ref{sec:cf-package}, we explain the overall structure and handling of the package as well as its most important functionalities. 
We present use cases for a regression and classification task to show the main functionalities of the package in Section~\ref{sec:use-cases}, followed by an example in Section~\ref{extension-of-the-package} illustrating how additional counterfactual explanation methods can be easily integrated into our package. 
In Section~\ref{benchmarking}, we show the general setup and results of the benchmark study.
%While new counterfactual explanation methods are continually being developed, there is a lack of benchmark studies comparing several methods. Exceptions are \cite{ref-pawelczyk2021carla}, \cite{ref-deoliveira21benchmark}, and \cite{ref-dandl2020}.
We summarize our findings as well as open questions in Section~\ref{conclusion}.

%% -- Manuscript ---------------------------------------------------------------

%% - In principle "as usual" again.
%% - When using equations (e.g., {equation}, {eqnarray}, {align}, etc.
%%   avoid empty lines before and after the equation (which would signal a new
%%   paragraph.
%% - When describing longer chunks of code that are _not_ meant for execution
%%   (e.g., a function synopsis or list of arguments), the environment {Code}
%%   is recommended. Alternatively, a plain {verbatim} can also be used.
%%   (For executed code see the next section.)

%% -- Methodological background -------------------------------------------------------------

\section{Methodological background and extensions} \label{sec:methods}

Our definition of counterfactual explanations is based on the work of \cite{ref-dandl2020} and \cite{ref-verma2020counterfactual}.

\begin{definition}[Counterfactual explanation]
 \label{def:cfe}

Let $\hat{f}:\mathcal{X} \rightarrow \mathbb{R}$ be a prediction function with $\mathcal{X} \subset \mathbb{R}^{p}$ as the feature
space. While our definition naturally covers regression models, for classification tasks, we assume that $\hat{f}$ returns the score or probability for a 
a predefined class of interest, usually the so-called positive class.
%For the multiclass case with $g$ classes, $\hat{f}(\mathbf{x})$ is usually assumed to return a class label or a vector of probabilities/scores $(\hat{f}_1(\mathbf{x}), ..., \hat{f}_g(\mathbf{x}))$. For convenience, we assume $\hat{f} := \hat{f}_k$ for a predefined class $k$ of interest.  
Let further $\mathbf{X} :=  (\mathbf{x}^{(1)}, ..., \mathbf{x}^{(n)})$ with  $\mathbf{x}^{(i)} \in \mathcal{X}, i \in \{1, ..., n\}$ be the observed data and $Y' = [Y_l', \; Y_u']$ be an
interval of desired predictions. We define a point $\mathbf{x}$ as a counterfactual explanation for an
observation $\mathbf{x}^{\star}$ if $\mathbf{x}$ fulfills (at least some of) the following desired
properties:

\begin{enumerate}[i]
\item \label{validity}
  \textit{Validity:} $\mathbf{x}$ leads to a desired prediction, i.e., $\hat{f}(\mathbf{x}) \in Y'$.
  This could be assessed, e.g., by \citep{ref-dandl2020}
 \begin{equation}
   o_{\text{valid}}\left(\hat{f}(\mathbf{x}), Y'\right)=\left\{\begin{array}{ll}0, & \text { if } \hat{f}(\mathbf{x}) \in Y' 
      \\ \min _{y' \in Y'}\left|\hat{f}(\mathbf{x})-y'\right|, & \text { otherwise }\end{array}.\right. 
    \label{eq:validity}
\end{equation}
\item \label{proximity}
  \textit{Proximity:} $\mathbf{x}$ is close to $\mathbf{x}^{\star}$, which could be measured, e.g., by the Gower distance $d_G$ \citep{ref-gower1971general} for mixed feature spaces 
  \begin{equation} 
      o_{\text{prox}}\left(\mathbf{x}, \mathbf{x}^{\star}\right)= d_G(\mathbf{x}, \mathbf{x}^{\star}) := \frac{1}{p} \sum_{j=1}^{p} \delta_{G}\left(x_{j}, x_{j}^{\star}\right) \in[0,1]
     \label{eq:gower}
    \end{equation} 
    with
    \begin{equation*} 
  \delta_G(x_j, x_j^{\star}) = \left\{\begin{array}{ll}\frac{1}{\hat{R}_j} \left|x_{j}-x_{j}^{\star}\right| & \text { if } x_{j} 
  \text { is numerical } \\ \mathbb{I}_{x_{j} \neq x_{j}^{\star}} & \text { if } x_{j} \text { is categorical }\end{array}.\right.
\end{equation*}
where $\hat{R}_j = \max(\mathbf{X_j}) - \min(\mathbf{X_j})$ is the value
range of feature $j$ in $\mathbf{X}$. 
\item \label{sparsity}
  \textit{Sparsity:} $\mathbf{x}$ differs from $\mathbf{x}^{\star}$ in only a few features. This can be measured by the $L_0$ norm
  \begin{equation} 
      o_{\text{sparse}}\left(\mathbf{x}, \mathbf{x}^{\star}\right)=\left\|\mathbf{x}-\mathbf{x}^{\star}\right\|_{0}=\sum_{j=1}^{p} 
      \mathbb{I}_{x_{j} \neq x_{j}^{\star}}.
      \label{eq:sparsity}
    \end{equation} 
\item \label{plausibility}
  \textit{Plausibility:} $\mathbf{x}$ is realistic, i.e., close to the data manifold. Metrics are the (weighted) Gower distance to the $k$ closest training samples $\mathbf{x}^{[1]}, ..., \mathbf{x}^{[k]} \in  \mathbf{X}$ \citep{ref-dandl2020}
  \begin{equation} 
      o_{\text{plaus}}\left(\mathbf{x}, \mathbf{X}\right)=\sum_{i=1}^{k} w^{[i]} d_{G}\left(\mathbf{x}^{[i]}, \mathbf{x}^{\star}\right) \in[0,1] \text { where } \sum_{i=1}^{k} w^{[i]}=1
      \label{eq:plausibility}
    \end{equation} 
  or the reconstruction error of a variational autoencoder (VAE) trained on the training samples \citep{ref-brughmans2021nice}. 
  \item \label{actionability}
  \textit{Actionability:} $\mathbf{x}$ does not alter immutable features (e.g., country of birth) and only proposes changes within an actionable range (e.g., non-negative age).
  \item \label{causality}
  \textit{Causality:} $\mathbf{x}$ reflects the underlying causal structure and takes causal relations of features into account. This property could be only examined if the causal graph \citep{ref-pearl2000dag} is (at least partially) known \citep{ref-karimi2020model, ref-karimi2021algorithmic, ref-mahajan2020preserving}. Since this is rarely the case, most counterfactual methods (including the ones implemented in the \pkg{counterfactuals} package) disregard this property \citep{ref-verma2020counterfactual}. 
\end{enumerate}

\end{definition}

While some desired properties have a common tendency, others are rather opposed: if an
explanation is sparse (\ref{sparsity}), it also tends to be proximal (\ref{proximity}), since a counterfactual tends to 
be close to the original data point when only a few features are changed. However, a counterfactual that is close to the original data point tends to have a similar prediction, which may be far from 
a desired prediction, thus making the counterfactual less valid (\ref{validity}). The exact interdependence between
the properties depends on the prevailing circumstances. 
Existing counterfactual methods vary in the desired properties they consider and how they measure and optimize them. An overview of methods is given in \cite{ref-verma2020counterfactual}. The methods also vary in whether a single counterfactual or a set of diverse ones is generated for a $\mathbf{x}^\star$. We argue that a set of counterfactuals is more valuable than a single one. This is because there could exist different equally good counterfactuals with the desired prediction (Rashomon effect \citep{ref-Breiman2001}) and it is more likely that a set contains a counterfactual that satisfies a user's (hidden) preferences \citep{ref-dandl2020}. 

Below, we introduce the three counterfactual methods currently available in the \pkg{counterfactuals} package: \textit{MOC} \citep{ref-dandl2020}, \textit{WhatIf} \citep{ref-wexler2019if}, and \textit{NICE} \citep{ref-brughmans2021nice}.
By addressing their limitations, we motivate \textit{optional} extensions of the methods that we implemented in our package.
In particular, these extensions enable all methods to return multiple counterfactuals for binary and multiclass classification models, as well as regression models.

%% -- MOC -------------------------------------------------------------

\subsection{Multi-objective counterfactual explanations}\label{MOC22}

\subsubsection{Original method}
The multi-objective counterfactuals (\textit{MOC}) method by \cite{ref-dandl2020} searches for counterfactuals by solving a multi-objective minimization problem

\begin{equation} 
  \min_{\mathbf{x}} \mathbf{o}(\mathbf{x}):=\min_{\mathbf{x}}\left(o_{\text{valid}}(\hat{f}(\mathbf{x}), Y'), o_{\text{prox}}\left(\mathbf{x}, \mathbf{x}^{\star}\right), o_{\text{sparse}}\left(\mathbf{x}, \mathbf{x}^{\star}\right), o_{\text{plaus}}(\mathbf{x}, \mathbf{X})\right).
  \label{eq:moc-min-prob}
\end{equation}

The single objectives correspond to the desired properties \emph{Validity}, \emph{Proximity}, \emph{Sparsity}, and \emph{Plausibility} formalized in Equations~\ref{eq:validity}~to~\ref{eq:plausibility} as part of Definition \ref{def:cfe}. 
\textit{MOC} also considers \emph{Actionability} by allowing the specification of ``fixed features'' that remain unchanged and of alteration ranges for continuous features.

To tackle the optimization problem in \eqref{eq:moc-min-prob}, \textit{MOC} uses a customized version of the non-dominated sorting genetic
algorithm (NSGA-II) of \cite{ref-deb2002fast}: unlike the original algorithm, \textit{MOC} employs mixed-integer
evolutionary strategies \citep{ref-li2013} to handle mixed feature spaces and computes the crowding distance not only in the objective space but also in the feature space.
A description of the steps of the algorithm as implemented in the \pkg{counterfactuals} package is given in Algorithm~\ref{algo:moc} of Appendix~\ref{app:algo-ref}.

The algorithm first initializes a population. The authors proposed several strategies: 
\begin{itemize}
\item
  \emph{Random}: Feature values of new individuals are uniformly sampled from the range of observed
  values. Subsequently, some  features are randomly reset to their initial value in
  $\mathbf{x}^{\star}$ to induce sparsity.
\item
  \emph{ICE curve}: As in \emph{Random}, feature values are sampled from the range of observed
  values. Then, however, features are reset with probabilities relative to their feature
  importance: the higher the importance of a feature $\mathbf{x}_j$, the higher the
  probability that its values differ from $\mathbf{x}_j^{\star}$. The 
  importance of one feature is measured using the standard deviation of its corresponding individual conditional expectation (ICE) curve \citep{ref-goldstein2015}. %, which show for one
%   feature and one observation how the prediction changes for varying values of the feature
%   c.p.. The greater the change in prediction, the higher the standard deviation of the ICE
%   curve $\sigma_j^{ICE}$. The following transformation then converts a standard deviation $\sigma_j^{ICE}$ into a probability:
%   \begin{equation} 
%     P(\mathbf{x}_j \neq \mathbf{x}^{\star}_j) = \frac{(\sigma_j^{ICE} - min(\sigma^{ICE})) * (p_{max} -
%     p_{min})}{max(\sigma^{ICE}) - min(\sigma^{ICE})} + p_{min}
%   \end{equation}
%   where $p_{min}$ and $p_{max}$ are further control parameters with default values $0.01$ 
%   and $0.99$, respectively, and $\sigma^{ICE} = (\sigma_1^{ICE}, \dots, \sigma_p^{ICE})$.
\item
  \emph{Standard deviation}: This method is similar to \emph{Random}, except that the sample ranges of numerical
  features are limited to one standard deviation from their value in $\mathbf{x}^{\star}$.
\item
  \emph{Training data:} Contrary to the other strategies, individuals are drawn from
  non-dominated previous observations in the dataset. If insufficient observations are
  available, the remaining individuals are initialized by random sampling. Subsequently, some
  features are randomly reset to their initial value in $\mathbf{x}^{\star}$ (as for
  \emph{Random}).
\end{itemize}

\cite{ref-dandl2020} discussed only the first two strategies in their paper, although the third and fourth strategies were also available in their implementation \citep{ref-dandl2020github}.
In subsequent generations, the algorithm recombines and mutates individuals of the population and their features with predefined probabilities so that the initial population evolves.
%For recombination, \citep{ref-dandl2020} use a simulated binary crossover recombinator for 
%numerical features \citep{ref-deb1995simulated} and a uniform crossover
%recombinator for all others \citep{ref-Syswerda1989}.
For mutation, the authors state two approaches: the first is to apply a scaled Gaussian mutator 
to numerical features and a uniform discrete mutator to categorical features \citep{ref-li2013}; 
the second approach aims to take feature distributions into account by sampling conditionally on the other feature values
using a transformation tree \citep{ref-hothorn2017transformation}. 

After recombination and mutation, some features are randomly reset to their initial value in
$\mathbf{x}^{\star}$ with prespecified probabilities to induce sparsity.
The recombination and mutation steps in the algorithm can be customized via multiple control
parameters. An overview is given in Appendix~\ref{default-values}.
To emphasize \emph{Validity} (\ref{validity}), individuals whose prediction exceeds a specified target distance
$\epsilon \in \mathbb{R}_{\ge 0}$ can be penalized using the approach of \cite{ref-deb2002fast}. 
% after the individuals are assigned to fronts $F_1, \dots, F_K$ by non-dominated sorting, 
% distance violators are ordered by their degree of violation and moved up into fronts $F_{K+1}, F_{K+2}, ...,$ 
% reducing their chance of survival.
\textit{MOC} terminates either after a prespecified number of generations or when the hypervolume (HV) indicator \citep{ref-zitzler1998multiobjective} of the objectives in \eqref{eq:moc-min-prob} does not improve for a prespecified number of consecutive generations. 
As counterfactuals, \textit{MOC} returns all (unique) non-dominated individuals across all generations.
%Once \textit{MOC} terminates after a prespecified number of generations, it returns all (unique) non-dominated individuals across all generations as counterfactuals.

Contrary to most other methods, \textit{MOC} is inherently applicable to both classification and regression tasks. Moreover, \textit{MOC} does not require the user to weigh the objectives \textit{a priori} and thus avoids the risk of arbitrarily affecting the solution set. 
Instead, it returns a Pareto set of counterfactuals so that the objectives can be weighted \textit{a posteriori}.

\subsubsection{Modifications}
We did not rely on the previous implementation of \textit{MOC} \citep{ref-dandl2020github} in the \pkg{counterfactuals} \proglang{R} package. Instead, we reimplemented an updated version of \textit{MOC}:
we replaced the NSGA-II implementation in \pkg{mosmafs} \citep{ref-mosmafs2020} with its extended and more versatile successor \pkg{miesmuschel} \citep{ref-miesmuschel2021}, and parameter spaces are now defined by the \pkg{paradox} package \citep{ref-paradox2021} instead of \pkg{ParamHelpers} \citep{ref-paramhelpers2020}. 

%%%%
%% -- WhatIf -------------------------------------------------------------
%%%

\subsection{WhatIf}\label{whatif}

\subsubsection{Original method}
\textit{WhatIf} is the counterfactual method for classification models proposed by \cite{ref-wexler2019if}
as part of the What-If Tool\footnote{https://pair-code.github.io/what-if-tool/}.
\cite{ref-wexler2019if} assume that the underlying model $\hat{h}:\mathcal{X}\rightarrow \mathcal{Y}$ predicts a class label and define the set of desired predictions $Y'$ as the set of all labels other than the current
one. 
As a counterfactual $\mathbf{x}'$ for an observation $\mathbf{x}^{\star}$, \textit{WhatIf} returns the
data point most similar to $\mathbf{x}^{\star}$ from previous observations
$\mathbf{\tilde{X}} = \{\mathbf{x} \in \mathbf{X}: \hat{h}(\mathbf{x}) \neq \hat{h}(\mathbf{x^{\star}}) \}$ whose predicted class is different from that of $\mathbf{x}^{\star}$.
This leads to the minimization problem:

\begin{equation} 
  \mathbf{x}' \in \underset{\mathbf{x} \in \mathbf{\tilde{X}}}{\mathrm{argmin}}\, d(\mathbf{x}, \mathbf{x^{\star}}).
  \label{eq:whatif-min-prob}
\end{equation}
The function $d$ is a slightly adapted version of the Gower distance (Equation~\ref{eq:gower}): for 
numerical features, the authors scale the distances with the standard deviations $\hat{\sigma}_j$; for categorical features, the feature distances are set equal ``to the probability that any two examples across the entire dataset would share the same
value for that feature'' if their values differ, and 0 otherwise \citep{ref-wexler2019if}. 
By definition, \textit{WhatIf} aims for valid (\ref{validity}), proximal (\ref{proximity}), and plausible (\ref{plausibility}) counterfactuals. 
\textit{WhatIf} often serves as a baseline method in benchmark studies \citep{ref-dandl2020, ref-schleich2021geco, ref-CarreiraPerpinan2021oblique} because it is easily implementable and adaptable.

\subsubsection{Modifications}
For better comparability with \textit{MOC}, we use the original Gower distance as the default for $d$ in the \pkg{counterfactuals} package.
We allow users to replace this with other dissimilarity measures (see Section~\ref{subsec:extend-dist}).
We also extended the method to work with $\hat{f}$ that returns the probability of a prespecified class of interest for classification tasks instead of a hard label classifier $\hat{h}$. 
This allows us to define the set of desired predictions $Y'$ as a probability interval $[Y_l', \; Y_u'] \subseteq [0,1]$.
% This enables a more advanced search with \textit{WhatIf} because the desired predictions can be specified in greater detail. 
%For hard classifiers, $Y'$ can be set to 0 or 1, which corresponds to the original approach of \cite{ref-wexler2019if}.  
Additionally, our approach makes \textit{WhatIf} applicable to regression tasks without further modifications. In this case, $Y'$ can simply be any real interval.
$\mathbf{\tilde{X}}$ is then redefined as $\mathbf{\tilde{X}} = \{\mathbf{x} \in \mathbf{X}: \hat{f}(\mathbf{x}) \in Y' \}$.

As argued in Section~\ref{sec:intro}, methods that can find multiple counterfactuals for a single observation are preferable. 
Therefore, we implemented an extended \textit{WhatIf} version that returns the $l \in \mathbb{N}$ closest data points of $\mathbf{\tilde{X}}$ to $\mathbf{x}^\star$ with the desired prediction. 
This is equivalent to minimizing the following objective instead of~\eqref{eq:whatif-min-prob}
\begin{equation} 
   \{\mathbf{x}'_1, \dots, \mathbf{x}'_l\} \in  \underset{\mathbf{Z} \subset \mathbf{\tilde{X}}, \; \left| 
   \mathbf{Z}\right| = l }{\mathrm{argmin}} \, \sum_{\mathbf{z} \in \mathbf{Z}} d_{G}(\textbf{z}, \textbf{x}^{\star}).
   \label{eq:ext-whatif-min-prob}
 \end{equation}
% 

%% -- NICE -------------------------------------------------------------
\subsection{Nearest instance counterfactual explanations}\label{nice-chapter}

\subsubsection{Original method}
Nearest instance counterfactual explanations (\textit{NICE}) introduced by \cite{ref-brughmans2021nice} is a counterfactual explanation method for binary score classifiers $\hat{f}: \mathcal{X} \rightarrow [-1,1]$. Accordingly, they define the set of desired predictions $Y'$ as the set of all scores that lead to a different class 
than the current one.
\textit{NICE} starts the counterfactual search for an observation $\mathbf{x}^{\star}$ by finding its most
similar \textit{correctly classified} instance $\mathbf{x}_{nn}$.
% Correctly classified means that $\hat{f}(\mathbf{x}_{nn}) = y_{nn}$ which could only be evaluated if $y_{nn}$ is known. 
\cite{ref-brughmans2021nice} assess similarity by the heterogeneous euclidean overlap method
\citep{ref-wilson1997improved} with $L_1$-norm aggregation, which corresponds to the Gower distance without averaging (i.e., Equation~\ref{eq:gower} without $\frac{1}{p})$.

Once $\mathbf{x}_{nn}$ is found, \textit{NICE} generates new instances in the first iteration ($m = 1$) by replacing single feature values of $\mathbf{x}^{\star}$ with the corresponding value of $\mathbf{x}_{nn}$.
\textit{NICE} evaluates the created instances with a reward function that optimizes either sparsity, proximity, or plausibility \citep[see][for details]{ref-brughmans2021nice}.

If the prediction of the instance with the highest reward 
value is in $Y'$, the algorithm terminates and returns this instance as a counterfactual.
Otherwise, \textit{NICE} creates new instances in the next iteration by replacing single feature values of the best performing instance of the previous iteration with the corresponding value of $\mathbf{x}_{nn}$.
The search continues as long as the prediction for the highest reward 
value instance is not in $Y'$. 

\subsubsection{Modifications}
We generalized \textit{NICE} for regression models and multiclass classifiers: first, we extend $\hat{f}$ to predict real-values (regression) or the probability of a predefined class $k$, respectively (see Definition~\ref{def:cfe}).
Second, we conceptualize the search for $\mathbf{x}_{nn}$ as the following minimization problem:

\begin{equation} 
  \mathbf{x}_{nn} = \underset{\mathbf{x} \in \mathbf{\mathring{X}'}}{\mathrm{argmin}} \, o_{\text{prox}}(\mathbf{x}, \mathbf{x}^*)
  \label{eq:nice_xnn}
\end{equation}

with $o_{\text{prox}}$ as defined in Equation~\ref{eq:gower}.  For classification, $\mathbf{\mathring{X}'} = \{\mathbf{x} \in \mathbf{X}: \hat{f}(\mathbf{x}) \in Y' \land h(\hat{f}(\mathbf{x})) = y \}$ is the set of all correctly classified observations whose prediction is in the set of desired
predictions $Y'$. $y$ is the true class label of $\mathbf{x}$ and $h(\cdot)$ is a
transformation function that maps class scores onto class labels.
For regression, $\mathbf{\mathring{X}'} = \{\mathbf{x} \in \mathbf{X}: \hat{f}(\mathbf{x}) \in Y' \land |\hat{f}(\mathbf{x}) - y| \le \epsilon \}$ is the set of all observations with a prediction in the desired real interval $Y'$ and a prediction error of less than a user-specified $\epsilon \in \mathbb{R}_{\ge 0}$.
Similar to \textit{WhatIf}, $o_{\text{prox}}$ in Equation~\ref{eq:nice_xnn} could be replaced with user-defined distance measures in our implementation (demonstrated in Section~\ref{subsec:extend-dist}).

The whole process after finding $\mathbf{x}_{nn}$ is already applicable to both multiclass classification and regression tasks.
We only updated the proposed reward functions for an iteration $m$ to 
\begin{equation} 
        R_O(\mathbf{x}) = \frac{o_{\text{valid}}(\hat{f}(\mathbf{x}_{m-1, R_{max}}), Y') - 
        o_{\text{valid}}(\hat{f}(\mathbf{x}), Y')}{O(\mathbf{x}, \mathbf{x}_{m-1, R_{max}} \mid \mathbf{x}^{\star})}
        \label{eq:reward},
\end{equation} 
 where $\mathbf{x}_{i-1, R_{max}}$ is the highest reward instance of the previous iteration ($m-1$), and $o_{\text{valid}}$ is defined in Equation~\ref{eq:validity}.
 The denominator $O(\cdot, \cdot)$ corresponds to the originally proposed functions aiming either at sparsity, proximity, or plausibility. 
 
Although multiple instances could have the desired prediction (and similar reward values), the original \textit{NICE} algorithm only returns a single counterfactual. 
In the \pkg{counterfactuals} package, we implemented two (optional) extensions that enable \textit{NICE}
to return multiple counterfactuals.
Our first extension returns all created instances (from all iterations) with a
desired prediction as counterfactuals after termination. 
Our second extension does not terminate when the prediction of
the highest reward instance is in the desired interval. Instead, it continues until $\mathbf{x}_{nn}$ is recreated. This
leads to a total number of $(d^2 + d) / {2}$ created instances, where $d$ is the number of feature values that differ
between $\mathbf{x}^{\star}$ and $\mathbf{x}_{nn}$. Like our first extension, it then returns all created instances with a desired prediction as counterfactuals. 
Compared to counterfactuals in earlier iterations, a counterfactual created in a later iteration is
inferior w.r.t.\ \emph{Proximity} (\ref{proximity}) and \emph{Sparsity} (\ref{sparsity}) (as more feature values are changed), but may be superior w.r.t.\ \emph{Plausibility} (\ref{plausibility}).
The pseudocode of our modified \textit{NICE} version is shown in Algorithm~\ref{algo:nice} of Appendix~\ref{app:algo-ref}.

In contrast to \textit{MOC}, \textit{NICE} does not consider all the desired counterfactual properties (listed in
Definition~\ref{def:cfe}) simultaneously: while \textit{NICE} guarantees \emph{Validity} by design (provided
that a correctly classified observation with a desired prediction exists), the user must
prioritize the other desired properties under the given circumstances and choose the reward
function accordingly. If there is no clear preference for the properties \textit{a priori}, we recommend
running our second \textit{NICE} extension for each of the reward functions, combining the counterfactuals, removing duplicates, and evaluating the remaining counterfactuals \textit{a posteriori}. 
We chose this strategy for our benchmark study in Section~\ref{benchmarking}. 

% Our two implemented extensions enable \textit{NICE} to return multiple counterfactuals. 
A not yet implemented extension is to set lower and upper bounds on $\mathbf{x}_{nn}$ to constrain the
feature values of the counterfactuals, enhancing their \emph{Actionability} \eqref{actionability}. 
Another extension would be to run the algorithm multiple times, defining $\mathbf{x}_{nn}$
in the $l\text{-th}$ run as the $l\text{-th}$ most similar (correctly classified) data point of
$\mathbf{x}^{\star}$, which increases the diversity of the counterfactuals.

%% -- The counterfactuals R package -------------------------------------------------------------

\section[The counterfactuals R package]{\pkg{counterfactuals} \proglang{R} package}\label{sec:cf-package}

In this section, we introduce the \pkg{counterfactuals} \proglang{R} package and explain its structure and workflow. The package is available from the Comprehensive \proglang{R} Archive Network (CRAN) \citep{ref-counterfactuals2022}. 

\begin{figure}
\centering
\includegraphics[width=0.8\linewidth,trim={0cm 6cm 0 0},clip]{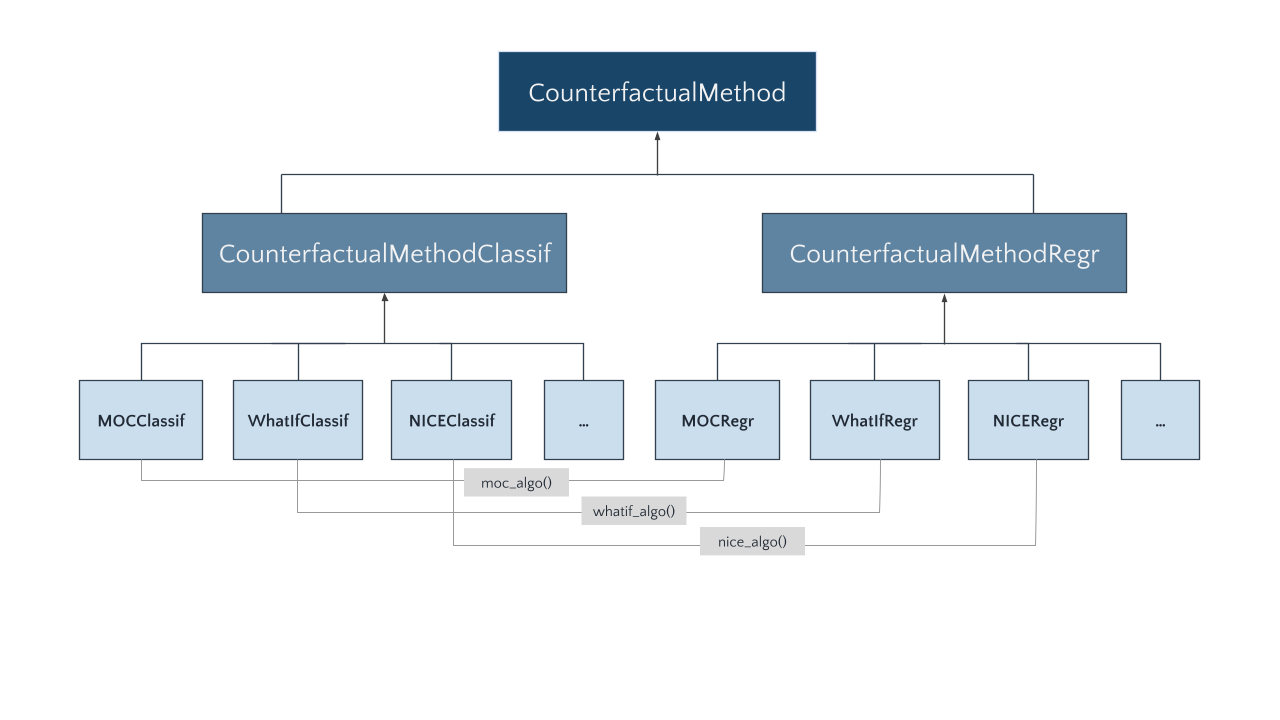}
\caption{Inheritance diagram of the \pkg{counterfactuals} package; a more detailed version is included in Appendix~\ref{class-diagram}.}
\label{fig:img-class-diagram}
\end{figure}

Inspired by the \pkg{iml} package \citep{ref-Molnar2018}, each counterfactual method described in the previous section is implemented in \code{R6} classes \citep{ref-r6chang}. 
Datasets and counterfactuals are represented as \pkg{data.table} objects \citep{ref-datatable2021} to allow efficient data manipulations and computations. 
Depending on whether a
counterfactual method supports classification or regression tasks, its class inherits from the (abstract) \code{R6} class 
\code{CounterfactualMethodClassif} or \code{CounterfactualMethodRegr} classes, respectively. Counterfactual
methods that support both tasks are split into two separate classes. Figure \ref{fig:img-class-diagram} illustrates the inheritance structure. For instance, as \textit{MOC} is applicable to classification and regression tasks, we implemented two classes: \code{MOCClassif} and \code{MOCRegr}. Both classes rely on the same (private) code base (\code{moc_algo()}) to generate counterfactuals to avoid code repetitions.  
\code{MOCClassif} inherits features from its superclass \code{CounterfactualMethodClassif}, while \code{MOCRegr} inherits from \code{CounterfactualMethodRegr}. Both of these superclasses in turn have the \code{CounterfactualMethod} as their superclass.

To generate counterfactuals for an arbitrary model with a specific counterfactual explanation method, the following steps are necessary: 
First, an \code{iml:::Predictor} object which encapsulates a fitted model and the underlying data must be initialized. The \code{Predictor} object is a wrapper for any machine learning model and ensures a unified interface and output for model predictions. 
It offers the necessary flexibility to generate counterfactuals for models fitted with a variety of popular machine learning interfaces (e.g., fitted with the \pkg{caret} \citep{ref-kuhn2021caret}, \pkg{mlr} \citep{ref-bischl2016mlr}, or \pkg{mlr3} packages \citep{ref-mlr32019}). We showcase this in the upcoming sections and Appendix~\ref{ap:mlinterfaces}.
The instantiated \code{Predictor} object serves as an input for the \code{predictor} field of the initialization method of the \code{WhatIfClassif/-Regr}, \code{MOCClassif/-Regr} or \code{NICEClassif/-Regr} classes. 
Additionally, the user can change the parameters of the used methods when initializing the object -- such as the mutation probability for \textit{MOC} or the used reward function for \textit{NICE}. Overviews of the parameters are given in Tables~\ref{tab:params-whatif}~-~\ref{tab:params-nice} in Appendix~\ref{default-values}. 

\begin{figure}
\centering
\includegraphics[width=0.65\linewidth]{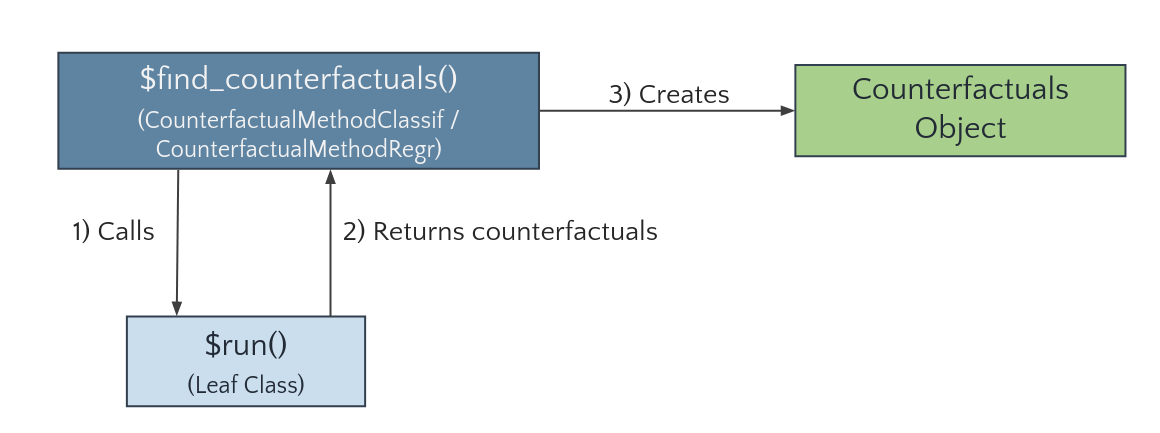} 

\caption{Call graph of the \pkg{counterfactuals} package. The \code{find\_counterfactuals()} method (1) calls a private \code{run()}
    method -- implemented by the leaf classes -- which performs the search and (2) returns the counterfactuals as a \code{data.table};
    \code{find\_counterfactuals()} then (3) creates a \code{Counterfactuals} object, which contains the counterfactuals and provides several
    methods for their evaluation and visualization.}\label{fig:img-call-diagram}
\end{figure}

Counterfactuals are generated by calling the \code{\$find\_counterfactuals()} method of the initialized object inherited from the classes \code{CounterfactualMethodClassif/-Regr}.
Figure~\ref{fig:img-call-diagram} illustrates the internal call graph.
As input, \code{find\_counterfactuals()} requires the observation of interest $\mathbf{x}^{\star}$ for which we seek counterfactuals as well as the desired prediction.
The method then calls the \code{\$run()} method, which is implemented in the leaf classes, and creates a \code{Counterfactuals} object that contains the generated counterfactuals. 
How the computational burden scales with the number of observations and number of features for the different methods is assessed in Section~\ref{benchmarking}.
Several tools are available to visualize and evaluate the counterfactuals.
They are showcased and explained in more detail in the upcoming section. 
These tools are primarily based on the codebase underlying \cite{ref-dandl2020}. More tools will be added in the future. 

%In addition, multiple components for generating counterfactuals can be exchanged, allowing methods to be easily extended and tailored to specific needs - for example, by adapting their dissimilarity measure (Section~\ref{subsec:extend-dist}).
%Due to the object-oriented concept of the package, users can also easily add their own counterfactual explanation methods. We provide an example in Section~\ref{extension-of-the-package}.

%% -- Use Cases -------------------------------------------------------------
\section{Use cases}\label{sec:use-cases}

In this section, we illustrate the \code{counterfactuals} workflow by applying \textit{MOC} (Section~\ref{MOC22}) to a classification task and our \textit{NICE} extension (Section~\ref{nice-chapter}) to a regression task. 
% We will also illustrate how the package can be extended with further counterfactual methods - Feature Tweaking \citep{ref-tolomei2017interpretable} as implemented in the \pkg{featureTweakR} package \citep{ref-kato2018github}.

\subsection{MOC applied to a classification task} \label{classification-task}

% To illustrate the \code{counterfactuals} workflow for classification tasks, we search for counterfactuals for
% diabetes tested-patients with \textit{MOC}.

As training data, we use the German Credit data set 
from the \pkg{rchallenge} package \citep{ref-rchallenge2021}.\footnote{The dataset was originally donated to UCI \citep{ref-uci2019} by Prof. Dr. Hofmann from Universit\"at Hamburg  and was later corrected by \cite{ref-groemping2019}.} The dataset originally contains 20 features on credit and personal information of 1000 bank customers. For illustrative purposes, we only consider the seven features: \code{duration, amount, purpose, age, employment\_duration, housing} and \code{number\_credits}. The target variable \code{credit\_risk} indicates whether a credit is a good/low or bad/high risk for the bank.

\begin{CodeChunk}
\begin{CodeInput}
R> library("counterfactuals")
R> library("iml")
R> library("randomForest")
R> data("german", package = "rchallenge")  
R> credit = german[, c("duration", "amount", "purpose", "age", 
+    "employment_duration", "housing", "number_credits", "credit_risk")]
\end{CodeInput}
\end{CodeChunk}

We train a random forest with the \pkg{randomForest} package to predict the \code{credit\_risk} \citep{ref-liaw2002randomForest}. We omit observation 998 from the training data, which is $\mathbf{x}^{\star}$, to imitate the situation of finding counterfactuals for a new observation.\footnote{This does not rule out the possibility to generate counterfactuals for training data points.}

\begin{CodeChunk}
\begin{CodeInput}
R> set.seed(20210816)
R> rf = randomForest(credit_risk ~ ., data = credit[-998L,])
\end{CodeInput}
\end{CodeChunk}

An \code{iml::Predictor} object serves as a wrapper for different model types. It contains the model and the data for its analysis.
We set \code{type = "prob"} such that class probabilities instead of hard labels are predicted. For our observation of interest $\mathbf{x}^{\star}$ -- denoted in the code as \code{x\_interest} -- the model predicts a probability of being a good credit risk  of 38.2\%:

\begin{CodeChunk}
\begin{CodeInput}
R> predictor = iml::Predictor$new(rf, type = "prob")
R> x_interest = credit[998L, ]
R> predictor$predict(x_interest)
\end{CodeInput}
\end{CodeChunk}

\begin{verbatim}
##     bad  good
## 1 0.618 0.382
\end{verbatim}

\subsubsection{Generation of counterfactuals}

Now, we examine which risk factors must be changed to increase the predicted probability of being a good credit risk to at least 60\%.
Since we want to apply \textit{MOC} to a classification model, we initialize a \code{MOCClassif} object. As explained in Section~\ref{MOC22}, individuals whose prediction is farther away from the desired interval than a prespecified value \code{epsilon} can be penalized. Here, we set \code{epsilon\ =\ 0} to penalize all individuals whose prediction is outside the desired interval. With the
\code{fixed\_features} argument, we fix the non-actionable features \code{age} and \code{employment\_duration} to the respective value of $\mathbf{x}^\star$.
By setting the termination criterion to \code{genstag}, we stop once the HV indicator does not increase for \code{n\_generations = 10L} consecutive generations.

\begin{CodeChunk}
\begin{CodeInput}
R> moc_classif = MOCClassif$new(
+   predictor, epsilon = 0, fixed_features = c("age", "employment_duration"),
+   termination_crit = "genstag", n_generations = 10L)
\end{CodeInput}
\end{CodeChunk}

We use the \code{\$find\_counterfactuals()} method to search for counterfactuals for \code{x\_interest}. As we aim to find
counterfactuals with a predicted probability of being a good credit risk of at least 60\%, we set the \code{desired\_class} to \code{"good"} and the
\code{predicted\_prob} to \code{c(0.6,\ 1)}; this is equivalent to setting the \code{desired\_class} to
\code{"bad"} and \code{desired\_prob} to \code{c(0,\ 0.4)}.

\begin{CodeChunk}
\begin{CodeInput}
R> cfactuals = moc_classif$find_counterfactuals(
+   x_interest, desired_class = "good", desired_prob = c(0.6, 1))
\end{CodeInput}
\end{CodeChunk}

\subsubsection[The Counterfactuals object]{The \code{Counterfactuals} object}
\label{subsubsec:cfactualsobj}

The resulting \code{Counterfactuals} object holds the counterfactuals in the \code{data} field and possesses several methods for their evaluation and visualization.
Printing a \code{Counterfactuals} object gives an overview of the results. Overall, we generated 82 counterfactuals.

\begin{CodeChunk}
\begin{CodeInput}
R> print(cfactuals))
\end{CodeInput}
\end{CodeChunk}

\begin{verbatim}
## 82 Counterfactual(s) 
##  
## Desired class: good 
## Desired predicted probability range: [0.6, 1] 
##  
## Head: 
##    duration amount purpose age employment_duration housing number_credits
## 1:       21   7460  others  30            >= 7 yrs     own              1
## 2:       21   7054  others  30            >= 7 yrs     own              1
## 3:       21   6435  others  30            >= 7 yrs     own              1
\end{verbatim}
The \code{\$predict()} method returns the predictions for the counterfactuals.

\begin{CodeChunk}
\begin{CodeInput}
R> head(cfactuals$predict(), 3L)
\end{CodeInput}
\end{CodeChunk}

\begin{verbatim}
##       bad  good
## 1:  0.322 0.678
## 2:  0.318 0.682
## 3:  0.296 0.704
\end{verbatim}

The \code{\$evaluate()} method returns the counterfactuals along with some predefined quality measures
\code{dist\_x\_interest}, \code{no\_changed}, \code{dist\_train}, and \code{dist\_target} 
for the desired properties \emph{Proximity}, \emph{Sparsity}, \emph{Plausibility}, and
\emph{Validity} (listed in Definition~\ref{def:cfe}). The quality measures are equal to the objectives of \textit{MOC}. Setting the \code{show\_diff} argument to \code{TRUE} displays 
the counterfactuals as their difference from \code{x\_interest}: for a numeric feature, positive
values indicate an increase compared to the feature value in \code{x\_interest} and negative 
values indicate a decrease; for factors, the feature value is displayed if it differs from
\code{x\_interest}; \code{NA} means ``no difference''.

\begin{CodeChunk}
\begin{CodeInput}
R> head(cfactuals$evaluate(show_diff = TRUE, measures = c("dist_x_interest", 
+     "dist_target", "no_changed", "dist_train")), 3L)
\end{CodeInput}
\end{CodeChunk}

\begin{verbatim}
##    duration amount purpose age employment_duration housing number_credits
## 1:       NA  -5220    <NA>  NA                <NA>    <NA>           <NA>
## 2:       NA  -5626    <NA>  NA                <NA>    <NA>           <NA>
## 3:       NA  -6245    <NA>  NA                <NA>    <NA>           <NA>
##    dist_x_interest no_changed dist_train dist_target
## 1:      0.04103193          1 0.04215022           0
## 2:      0.04422330          1 0.03895885           0
## 3:      0.04908897          1 0.03409318           0
\end{verbatim}

By design, there is no guarantee that all counterfactuals generated with MOC have a prediction $\in Y'$. Therefore, we use the \code{\$subset\_to\_valid()} method to omit all non-valid counterfactuals.
The method \code{\$revert\_subset\_to\_valid()} can reverse this step.

\begin{CodeChunk}
\begin{CodeInput}
R> cfactuals$subset_to_valid()
R> nrow(cfactuals$data)
## [1] 40 
\end{CodeInput}
\end{CodeChunk}

\begin{figure}[H]
    \centering
    \includegraphics{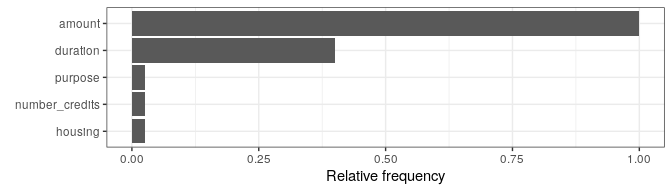}
    \caption{Relative frequency of feature changes across all counterfactuals. Features without proposed changes are omitted.}
    \label{fig:my_label}
\end{figure}

Of the 82 counterfactuals, 40 have the desired predictions.
To detect which features are the most important levers to obtain a certain prediction, the relative frequency of feature changes across all counterfactuals can be plotted via the \code{\$plot\_freq\_of\_feature\_changes()} method.
Setting \code{subset\_zero\ =\ TRUE} excludes all unchanged features from the plot. Figure~\ref{fig:my_label} shows that all counterfactuals require changes in the credit amount.

\begin{CodeChunk}
\begin{CodeInput}
R> cfactuals$plot_freq_of_feature_changes(subset_zero = TRUE)
\end{CodeInput}
\end{CodeChunk}

\begin{figure}[h]
    \centering
    \includegraphics{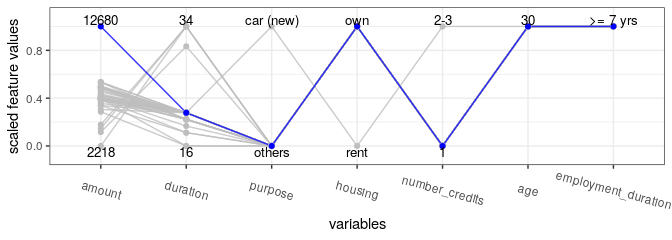}
    \caption{Parallel plot along (standardized) features. The blue line represents $\mathbf{x^\star}$ (\code{x\_interest}), whereas gray lines represent generated counterfactuals. 
    }
    \label{fig:credit-parallel}
\end{figure}

The parallel plot (Figure~\ref{fig:credit-parallel}) -- created with the \code{\$plot\_parallel()} method -- compares the feature values of the counterfactuals among each other (one gray line per counterfactual) and 
with \code{x\_interest} (blue line). Equal to \cite{ref-dandl2020}, all features are scaled between 0 and 1. 
The argument \code{feature\_names} filters the features and orders them, \code{NULL} means ``all''. Using \code{$get\_freq\_of\_feature\_changes()}, we order the features according to their frequency of changes.
The \code{digits\_min\_max} argument specifies the maximum number of digits for plotted values. The default value is \code{2L}.
All counteractuals propose a decrease in the credit \code{amount} while the \code{duration} either needs no modifications, an increase or an decrease. For one counterfactual, additionally the \code{purpose} was set to a new car, the \code{housing} type was set to rented and the \code{number\_credits} was increased.
\begin{CodeChunk}
\begin{CodeInput}
R> cfactuals$plot_parallel(feature_names = names(
+    cfactuals$get_freq_of_feature_changes()),  digits_min_max = 2L)
\end{CodeInput}
\end{CodeChunk}

\begin{figure}[H]
    \centering
    \includegraphics{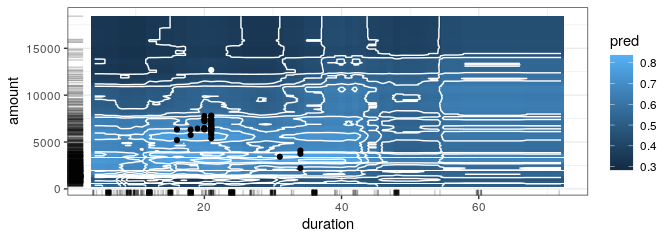}
    \caption{Prediction surface plotted along features \code{duration} and \code{amount}. Other feature values are held constant at $\mathbf{x^\star}$.
    % Predicted values are indicated by the colors and contour lines.
    The white point displays $\mathbf{x^\star}$. Black points are counterfactuals with variations only in the two displayed features.
    Rugs represent marginal distributions of the observed data.
    }
    \label{fig:credit-surface}
\end{figure}

The \code{\$plot_surface()} method generates prediction surface plots/2-dimensional ICE plots \citep{ref-dandl2020}. 
The method requires the names of two features (argument \code{feature_names}) as an input.
The white dot in Figure \ref{fig:credit-surface} represents \code{x\_interest}.
All counterfactuals that differ from \code{x\_interest} \textit{only} in the two selected features (here, \code{duration} and \code{amount}) are displayed as black dots.
We observe that either a change in \code{amount} alone, or in \code{amount} \textit{and} the \code{duration} is advocated. The rug lines next to the axes indicate the marginal distribution of the training data. It should be noted that the multi-objective approach does not consider counterfactuals farther away from \code{x_interest} as suboptimal because these counterfactuals outperform others in their proximity to the observed data points (plausibility property~(\ref{plausibility})).

\begin{CodeChunk}
\begin{CodeInput}
R> cfactuals$plot_surface(feature_names = c("duration", "amount"))
\end{CodeInput}
\end{CodeChunk}

\subsubsection{MOC diagnostics}

\begin{figure}[H]
    \centering
    \includegraphics{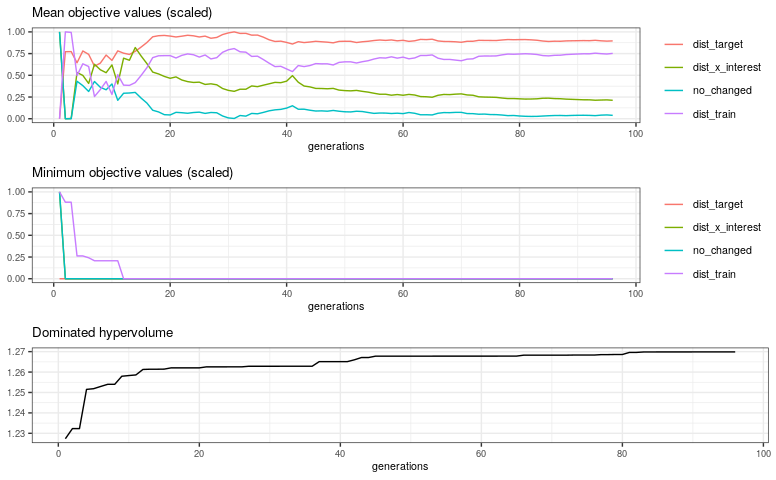}
    \caption{Evolution of the mean and minimum objective values together with the dominated HV over the generations. The mean and minimum objective values were scaled between 0 and 1.}
    \label{fig:credit-statistics}
\end{figure}

The aforementioned plotting and evaluation methods are part of the class \code{Counterfactuals} and all counterfactuals created by \textit{MOC}, \textit{WhatIf}, or \textit{NICE} can be evaluated with them. 
For \textit{MOC}, additional diagnostic tools are available. Since they are only applicable to \textit{MOC}, they cannot be called by the \code{Counterfactuals} class but rather by instances from the \code{MOCClassif} and \code{MOCRegr} class after counterfactuals were generated.  
To evaluate the estimated Pareto front, \cite{ref-dandl2020} use a HV indicator
\citep{ref-zitzler1998multiobjective} with reference point 
$s = (\text{inf}_{y' \in Y'} \; |f(\mathbf{x}^{\star}) - y'|, 1, p, 1)$ representing the maximal values of
the objectives ($o_{\text{valid}}$, $o_{\text{prox}}$, $o_{\text{sparse}}$, $o_{\text{plaus}}$ of Equations~\ref{eq:validity} to \ref{eq:plausibility}).
The evolution of the HV indicator can be plotted together with the evolution of mean and minimum 
objective values using the \code{\$plot\_statistics()} method. The \code{centered\_obj} argument allows the 
user to control whether the objective values should be centered: if set to \code{FALSE}, each objective value 
is visualized in a separate plot, since they (usually) have different scales; if set to \code{TRUE} (default),
they are visualized in a single plot, as shown in Figure~\ref{fig:credit-statistics}.

\begin{CodeChunk}
\begin{CodeInput}
R> moc_classif$plot_statistics(centered_obj = TRUE)
\end{CodeInput}
\end{CodeChunk}

Ideally, the mean value of each objective decreases, while the HV increases over the generations. However, there is often a trade-off between the objectives in the sense that when the mean value of one objective slightly decreases, it might slightly increase for another objective. 
This trade-off is also visible in the scatter plot created with the \code{\$plot\_search()}
method that visualizes the values of two specified \code{objectives} of all emerged individuals. Ideally, one 
would like to have a point shift to the lower-left corner over the generations, which implies lower and thus better objective values.

\begin{CodeChunk}
\begin{CodeInput}
R> moc_classif$plot_search(objectives = c("dist_train", "dist_target"))
\end{CodeInput}
\end{CodeChunk}

\begin{figure}[H]
    \centering
    \includegraphics{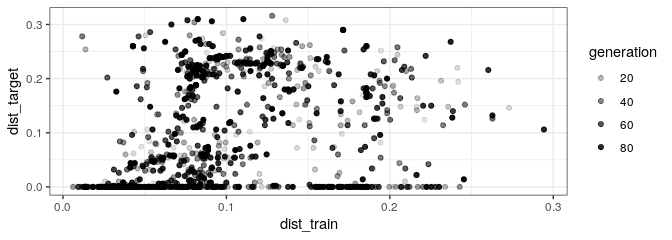}
    \caption{Evolution of the objectives \code{dist\_train} and \code{dist\_target} over the generations.}
    \label{fig:credit-search1}
\end{figure}

\begin{figure}[h!]
    \centering
    \includegraphics{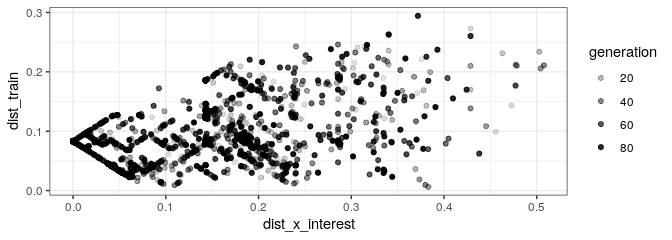}
    \caption{Evolution of the objectives \code{dist\_x\_interest} and \code{dist\_train} over the generations.}
    \label{fig:credit-search2}
\end{figure}

According to Figure~\ref{fig:credit-search1}, many counterfactual have predictions in the desired prediction range (\code{dist_target} = 0). However, many points for the objectives \code{dist\_train} and \code{dist\_target} are also located in the middle region. This underlines the difficulty of minimizing both objectives simultaneously. 
For the objectives \code{dist\_train} and \code{dist\_x\_interest} (Figure~\ref{fig:credit-search2}) (Figure~\ref{fig:credit-search2}), on the other hand, there is a clearer shift to the lower-left corner over the generations. The distinct boundary on the lower left indicates that the optimization potential for these two objectives might be fully exploited.

\begin{CodeChunk}
\begin{CodeInput}
R> moc_classif$plot_search(objectives = c("dist_x_interest", "dist_train"))
\end{CodeInput}
\end{CodeChunk}

\subsection{NICE applied to a regression task} \label{subsec:regression-task}

Searching for counterfactuals for regression models works analogously to classification models. In this example, we use our \textit{NICE} extension for regression models to search for multiple counterfactuals for a predictor of plasma retinol concentration. This is interesting because low concentrations are associated with an increased risk for some types of cancer (see \cite{ref-xie_etal_2019} for an overview).

As training data, we use the plasma dataset \citep{ref-harrison1978hedonic} from the \pkg{gamlss.data}
package \citep{ref-gamlss2021}. The dataset contains 315 observations with 13 features describing personal and dietary factors (e.g., age, number of alcoholic drinks per week or the measured plasma beta-carotene level) and the (continuous) target variable \code{retplasma} -- the plasma retinol concentration in ng/ml.
We train a regression tree with the \pkg{mlr3} package to predict \code{retplasma} \citep{ref-mlr32019}. 
We reserve the 100th row of the data for $\mathbf{x}^\star$ -- denoted as \code{x\_interest}. 

\begin{CodeChunk}
\begin{CodeInput}
R> library("mlr3") 
R> data("plasma", package = "gamlss.data")
R> x_interest = plasma[100L,]
R> tsk = mlr3::TaskRegr$new(id = "plasma", backend = plasma[-100L,], 
+    target = "retplasma")
R> tree = lrn("regr.rpart")
R> model = tree$train(tsk)
\end{CodeInput}
\end{CodeChunk}

Then, we initialize an \code{iml::Predictor} object. 
For \code{x\_interest}, the model predicts a plasma concentration of 342.92 ng/ml.
\begin{CodeChunk}
\begin{CodeInput}
R> predictor = Predictor$new(model, data = plasma, y = "retplasma")
R> predictor$predict(x_interest)
\end{CodeInput}
\end{CodeChunk}

\begin{verbatim}
##   pred
## 1 342.92
\end{verbatim}

Since we want to apply \textit{NICE} to a regression model, we initialize a \code{NICERegr} object.
The initial version of \textit{NICE} restricted to classification models starts the search by finding the most similar correctly classified datapoint. For regression models, we define a correctly predicted datapoint when its prediction is less than a user-specified value (\code{margin_correct}) away from the true outcome. In this example, we allow for a deviation of $0.5$.
The argument \code{optimization} specifies the reward function we want to optimize. We aim for the most proximal counterfactual by setting this argument to \code{proximal} and by setting \code{return_multiple} to \code{FALSE}.

We call the \code{\$find\_counterfactuals()} method to search for counterfactuals for
\code{x\_interest} with a predicted concentration of more than 500 ng/ml, i.e.\ a concentration in the interval $[500, Inf]$. 

\begin{CodeChunk}
\begin{CodeInput}
R> nice_regr = NICERegr$new(predictor, optimization = "proximity", 
+   margin_correct = 0.5, return_multiple = FALSE)
R> cfactuals = nice_regr$find_counterfactuals(x_interest, 
+   desired_outcome = c(500, Inf))
\end{CodeInput}
\end{CodeChunk}

\begin{figure}[H]
    \centering
    \includegraphics{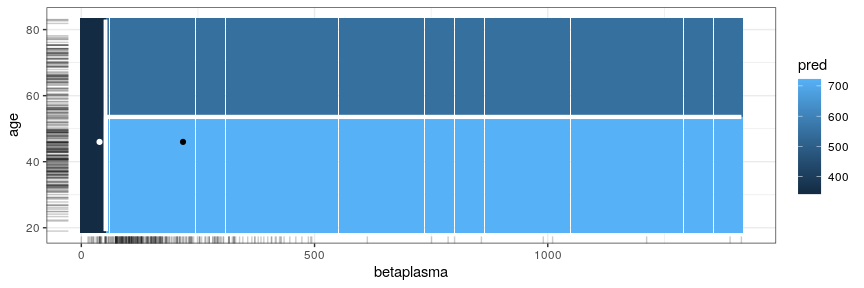}
    \caption{Prediction surface plotted along features \code{betaplasma} and \code{age}. Other feature values are held constant at $\mathbf{x^\star}$.
    % Predicted values are indicated by the colors and contour lines.
    The white point displays $\mathbf{x^\star}$. Black points are counterfactuals with variations only in the two displayed features.
    Rugs represent marginal distributions of the observed data. White horizontal lines are plotting artifacts.
    }
    \label{fig:plasma-surface}
\end{figure}

The result is a \code{Counterfactuals} object, which we can analyze with the same methods as in Section~\ref{subsubsec:cfactualsobj}.
The surface plot of plasma beta-carotene (\code{betaplasma}) and age (Figure~\ref{fig:plasma-surface}), for example, reveals that increasing the beta-carotene concentration (e.g., by eating more kale, carrots, etc.) is sufficient for predicting a plasma concentration $\ge 500$ ng/ml for $\mathbf{x}^\star$, while changing the age alone has no effect on the prediction.

\begin{CodeChunk}
\begin{CodeInput}
R> cfactuals$plot_surface(feature_names = c("betaplasma", "age"), grid_size = 200)
\end{CodeInput}
\end{CodeChunk}

\subsubsection{User-defined distance function}
\label{subsec:extend-dist}

As stated in Equation~\ref{eq:nice_xnn}, \textit{NICE} determines the most similar (correctly classified) datapoint by minimizing the Gower distance. 
However, the input parameter \code{distance_measure} of the initialization method of \code{NICERegr} (and \code{NICEClassif}) allows a different distance measure.
The parameter requires a function with arguments \code{x}, \code{y}, and \code{data}, that returns a numeric matrix with number of rows and columns corresponding to the number of observations in \code{x} and \code{y}, respectively.
As an example, we replace the Gower function with the $L_0$ norm. First, we set up the function and illustrate its functionality in a short example.
\begin{CodeChunk}
\begin{CodeInput}
R> l0_norm = function(x, y, data) {
+   res = matrix(NA, nrow = nrow(x), ncol = nrow(y))
+   for (i in seq_len(nrow(x))) {
+     for (j in seq_len(nrow(y))) {
+       res[i, j] = sum(x[i,] != y[j,])
+     }
+   }
+   res
+ }
R> xt = data.frame(a = c(0.5), b = c("a"))
R> yt = data.frame(a = c(0.5, 3.2, 0.1), b = c("a", "b", "a"))
R> l0_norm(xt, yt, data = NULL)
\end{CodeInput}
\end{CodeChunk}
\begin{verbatim}
##      [,1] [,2] [,3]
## [1,]    0    2    1
\end{verbatim}

Next, we forward this function to the \code{distance_function} argument of \code{NICERegr}. 
\begin{CodeChunk}
\begin{CodeInput}
R> nice_regr = NICERegr$new(predictor, optimization = "proximity", 
+   margin_correct = 0.5, return_multiple = FALSE, 
+   distance_function = l0_norm)
R> nice_regr$find_counterfactuals(x_interest, desired_outcome = c(500, Inf))
\end{CodeInput}
\end{CodeChunk}

\begin{verbatim}
## 1 Counterfactual(s) 
##  
## Desired outcome range: [500, Inf] 
##  
## Head: 
##    age sex smokstat    bmi vituse calories   fat fiber alcohol cholesterol 
## 1:  46   1        3  35.26      3   2667.5 131.6  10.1       0       550.5     
##    betadiet retdiet betaplasma
## 1:     1210    1291        218
\end{verbatim}

The initialization methods of \textit{MOC} and \textit{WhatIf} also have a \code{distance_function} argument: for \textit{MOC}, its input replaces the Gower distances used for $o_{\text{prox}}$ and $o_{\text{plaus}}$ (Equations~\ref{eq:gower}~\&~\ref{eq:plausibility}); for \textit{WhatIf}, its input replaces the Gower distance in Equation~\ref{eq:ext-whatif-min-prob}.

%% -- Extension of the package -------------------------------------------------------------
\section{Extension of the package}\label{extension-of-the-package}

We have designed the \pkg{counterfactuals} package 
to be quickly extensible by new methods. Here, we illustrate how to add new methods to the package by integrating the
\pkg{featureTweakR} package \citep{ref-kato2018github}, which implements Feature Tweaking
\citep{ref-tolomei2017interpretable}, a counterfactual method that can be applied to (classification) tree ensembles fitted with the \pkg{randomForest} package.
Feature Tweaking starts the search for counterfactuals for an observation $\mathbf{x}^{\star}$ by finding all trees in the
ensemble that do not predict the desired class. For each of these trees, it attempts to change (or ``tweak'') $\mathbf{x}^{\star}$
as little as possible to switch the prediction of that tree to the desired class. From all tweaked instances that also
switch the ensemble prediction to the desired class, it returns the tweaked instance
that changes $\mathbf{x}^{\star}$ the least as a counterfactual.

The \pkg{featureTweakR} package has a couple of limitations, e.g., factors in the training data cause problems or
that it is only applicable to random forests trained on standardized features with the \pkg{randomForest} package \citep{ref-liaw2002randomForest}. Due to these limitations, \pkg{featureTweakR}
is not part of the \pkg{counterfactuals} package but does serve as a suitable example here.
First, we install \pkg{featureTweakR} and its dependency \pkg{pforeach} \citep{ref-pforeach2015} and
load the required libraries.

\begin{CodeChunk}
\begin{CodeInput}
R> devtools::install_github("katokohaku/featureTweakR")
R> devtools::install_github("hoxo-m/pforeach")
R> library("featureTweakR")
R> library("counterfactuals")
R> library("iml")
R> library("randomForest")
R> library("R6")
\end{CodeInput}
\end{CodeChunk}

\subsection{Class structure}

At least two methods must be implemented for a new class: \code{\$initialize()} and \code{\$run()}.
The \code{\$print\_parameters()} method is not mandatory but still strongly recommended, as it gives objects of that class an
informative \code{print()} output.
As elaborated above, a new class inherits from either \code{CounterfactualMethodClassif} or \code{CounterfactualMethodRegr}, depending
on which task it supports. Since Feature Tweaking supports classification tasks, the new \code{FeatureTweakerClassif} class inherits
from the former.

\begin{CodeChunk}
\begin{CodeInput}
R> FeatureTweakerClassif = R6::R6Class("FeatureTweakerClassif", 
+   inherit = CounterfactualMethodClassif,
+   public = list(
+       initialize = function() {
+           # **see below**
+       }
+   ),
+   private = list(
+       run = function() {
+           # **see below**
+       },
+       print_parameters = function() {
+           # **see below**
+       }
+   )
+ )
\end{CodeInput}
\end{CodeChunk}

\subsubsection[Implementation of the initialize() method]{Implementation of the \code{\$initialize()} method}

In the next step, we implement the \code{$initialize()} method, which must have a \code{predictor} argument that takes an
\code{iml::Predictor} object. In addition, it may have further arguments specific to the counterfactual method. Feature Tweaking has the following hyperparameters:
\code{ktree} representing the number of trees to be considered, \code{epsiron}\footnote{Please note that this is not a typo on our part, but the naming in the original implementation \citep{ref-kato2018github}.} as the upper threshold of feature changes, and \code{resample} indicating whether trees are randomly selected or not.

\begin{CodeChunk}
\begin{CodeInput}
R> initialize = function(predictor, ktree = NULL, epsiron = 0.1, 
+   resample = FALSE) {
+    # adds predictor to private$predictor field
+    super$initialize(predictor) 
+    private$ktree = ktree
+    private$epsiron = epsiron
+    private$resample = resample
+  }
\end{CodeInput}
\end{CodeChunk}

We also fill the \code{\$print\_parameters()} method with the parameters of Feature Tweaking.

\begin{CodeChunk}
\begin{CodeInput}
R> print_parameters = function() {
+    cat(" - epsiron: ", private$epsiron, "\n")
+    cat(" - ktree: ", private$ktree, "\n")
+    cat(" - resample: ", private$resample)
+     }
\end{CodeInput}
\end{CodeChunk}

\subsubsection[Implementation of the run method]{Implementation of the \code{\$run()} method}

The \code{\$run()} method performs the search for counterfactuals. Its structure is completely free, which makes it flexible to add new
counterfactual methods to the \pkg{counterfactuals} package. The only requirement is that a \code{data.table} with the generated counterfactuals is returned at the end. The columns display the features and rows the counterfactuals.
%\footnote{Additional information could be stored in object properties and later retrieved via the \code{data} field of the resulting \code{Counterfactuals} object.}

The \code{\$run()} method is called by the method \code{\$find_counterfactuals()} implemented in the \code{CounterfactualMethodsClassif} class. As shown in Section~\ref{classification-task}, \code{\$find_counterfactuals} requires as input \code{x_interest}, \code{desired_class}, and \code{desired_prob}, which are saved in private fields. Thus, \code{\$run()} could directly access the information and preprocesses them before it passes them on to the implemented methods of \pkg{featureTweakR}.

The workflow of finding counterfactuals for \code{x_interest} with the \pkg{featureTweakR} package for a fitted random forest model \code{rf} consists of three steps: First, decision trees are transformed to data frames of paths by \code{getRules()}. Then, \code{set.eSatisfactory()} generates new instances by slightly altering feature values. Finally, \code{tweak()} generates counterfactuals for a specific instance $\mathbf{x}^\star$. Further information could be found in the documentation of the package \citep{ref-kato2018github}. 
The \code{\$run()} method encapsulates these steps and returns a data.frame of generated counterfactuals. 

\begin{CodeChunk}
\begin{CodeInput}
R> run = function() {
+    # Extract info from private fields
+    predictor = private$predictor
+    y_hat_interest = predictor$predict(private$x_interest)
+    class_x_interest = names(y_hat_interest)[which.max(y_hat_interest)]
+    rf = predictor$model
+    # Call functions in featureTweakR 
+    rules = getRules(rf, ktree = private$ktree, resample = private$resample)
+    es = set.eSatisfactory(rules, epsiron = private$epsiron)
+    tweaks = tweak(
+      es, rf, private$x_interest, label.from = class_x_interest, 
+      label.to = private$desired_class, .dopar = FALSE
+    )
+    return(tweaks$suggest)
+  }
\end{CodeInput}
\end{CodeChunk}
The composite code of our new class can be seen in Appendix~\ref{ap:featuretweak}. 

\subsection{Feature Tweaking applied to a classification task}

For demonstration purposes, we apply the implemented Feature Tweaking to the \code{iris} dataset \citep{ref-fisher1936iris, ref-anderson1936-iris}. 
We train a random forest on the dataset and set up the \code{iml::Predictor} object, again omitting \code{x\_interest} (here, row 130) from the training data.

\begin{CodeChunk}
\begin{CodeInput}
R> set.seed(78546)
R> X = subset(iris, select = -Species)[-130L,]
R> y = iris$Species[-130L]
R> rf = randomForest(X, y, ntree = 20L)
R> predictor = iml::Predictor$new(rf, data = iris[-130L, ], 
+   y = "Species", type = "prob")
\end{CodeInput}
\end{CodeChunk}

For \code{x\_interest}, the model predicts a probability of 30\% for \code{versicolor}.

\begin{CodeChunk}
\begin{CodeInput}
R> x_interest = iris[130L, ]
R> predictor$predict(x_interest)
\end{CodeInput}
\end{CodeChunk}

\begin{verbatim}
##   setosa versicolor virginica
## 1      0        0.3       0.7
\end{verbatim}

Now, we use Feature Tweaking to address the question: ``What changes in \code{x\_interest} are necessary for the model to predict
a probability of at least 60\% for \code{versicolor}?''$.$

\begin{CodeChunk}
\begin{CodeInput}
R> # Set up FeatureTweakerClassif
R> ft_classif = FeatureTweakerClassif$new(predictor, ktree = 10L, 
+   resample = TRUE)
R> # Find counterfactuals and create a Counterfactuals object
R> cfactuals = ft_classif$find_counterfactuals(
+   x_interest, desired_class = "versicolor", desired_prob = c(0.6, 1)
+  )
\end{CodeInput}
\end{CodeChunk}

As for \textit{MOC} and \textit{NICE}, the result is a \code{Counterfactuals} object which could be visualized and evaluated as shown in Section~\ref{subsubsec:cfactualsobj}.

% \begin{CodeChunk}
% \begin{CodeInput}
% R> cfactuals
% \end{CodeInput}
% \end{CodeChunk}

% \begin{verbatim}
% ## 1 Counterfactual(s) 
% ## 
% ## Desired class: versicolor 
% ## Desired predicted probability range: [0.6, 1] 
% ## 
% ## Head: 
% ##    Sepal.Length Sepal.Width Petal.Length Petal.Width
% ## 1:          7.2           3         4.85        1.55
% \end{verbatim}

\normalsize

% A minor limitation of this basic implementation is that one would not be able to find counterfactuals for a setting
% with \code{max(desired\_prob)\ \textless{}\ 0.5}, since \code{featureTweakR::tweak()} only searches for instances that would be predicted as
% \code{desired\_class} by majority vote. To enable this setting, one would need to change some \pkg{featureTweakR} internal code,
% but for the sake of clarity we will refrain from this here.

%% -- Benchmarking ------------------------------------------------------------
\section{Benchmarking}\label{benchmarking}

In this section, we use a benchmark study to answer the following research questions: 
\begin{enumerate}
    \item How do the different methods implemented in the \pkg{counterfactuals} \proglang{R} package perform according to the properties validity \eqref{validity}, proximity \eqref{proximity}, sparsity \eqref{sparsity} and plausibility \eqref{plausibility}  of Definition~\ref{def:cfe}, and according to the HV indicator and number of non-dominated counterfactuals? \label{RQ:1}
    \item How do the methods differ in their runtime for an increasing number of observations ($n$) and number of features ($p$)? \label{RQ:2}
\end{enumerate}
The overall design of our benchmark study is strongly inspired by the work of \cite{ref-dandl2020} who also compared different methods according to the four properties of Definition~\ref{def:cfe}. Aditionally, we evaluate the methods with regard to their runtime behavior and HV. Furthermore, we added \textit{NICE} as another comparison method. Since our source code is openly available\footnote{\url{https://github.com/slds-lmu/benchmark_2022_counterfactuals}}, we encourage readers to add other counterfactual methods to our \proglang{R} package and to compare them to the already implemented ones using our study code.

\newpage
\subsection{Setup}\label{setup}
We used six datasets from the OpenML platform \citep{ref-Vanschoren2014} 
with binary classes, no missing values, and varying numbers of observations and features. Table
\ref{tab:benchmarkingdatasets} provides an overview of the datasets.
To study the runtime behavior, we also ran all available methods on row-wise subsets (with differing number of observations $n \in \{886 \; (1\%),\, 8859 \; (10\%), \, 88588 \; (100\%)\}$) of the \code{run\_or\_walk\_information} dataset and column-wise subsets (with differing number of features
$p \in \{10, \, 30, \, 100\}$) of the \code{hill\_valley} dataset. The subsets were randomly generated and identical for all models and methods.

\begin{table}[t!]
\centering
\begin{tabular}{lllllp{7.4cm}}
\hline
OpenML ID & Name & Obs & Cont & Cat \\ \hline
31 & credit\_g & 1,000 & 7 & 13\\
37 & diabetes & 768 & 8 & 0\\
50 & tic\_tac\_toe & 958 & 0 & 9\\
725 & bank8FM & 8,192 & 8 & 0\\
1479 & hill\_valley & 1,212 & 100 & 0\\
40922 & run\_or\_walk\_information & 88,588 & 6 & 0\\ \hline
\end{tabular}
\caption{\label{tab:benchmarkingdatasets}Description of the OpenML datasets used for benchmarking. Obs displays the no. of observations, Cont the no. of continuous features and Cat the no. of categorical features.}
\end{table}

On each dataset, we tuned and trained five models using the \pkg{mlr3} \proglang{R} package \citep{ref-mlr32019}: a random forest (ranger), an xgboost, an RBF support vector machine (svm), a logistic regression (logreg), and a neural network with one hidden
layer (neuralnet).\footnote{For the \textbf{hill\_valley} dataset with 100 features, two dense layers were necessary.}  Beforehand, we standardized numerical features and one-hot-encoded categorical ones. For tuning, we employed random search with 30 evaluations and 5-fold cross-validation (CV) using the misclassification error as a performance measure.
Further details on the tuning search space and the classification accuracies are given in Appendix~\ref{benchmarking-section}.  Before training, 
we randomly selected ten observations from each dataset as $\mathbf{x}^{\star}$ and omitted them from the training
data. For each $\mathbf{x}^{\star}$, we set the desired class probability interval $Y'$ to the
opposite of the predicted class (based on a threshold of 0.5):

\begin{equation} 
  Y' = \left\{\begin{array}{ll} ]0.5, 1]  & \text { if } f(\mathbf{x}^{\star}) \leq 0.5 \\ 
  \left[0, 0.5 \right] & \text { else } \end{array}\right. .
\end{equation}

For each dataset, model, and $\mathbf{x}^{\star}$, we computed counterfactuals with \textit{WhatIf}, \textit{NICE} and \textit{MOC}. 
Apart from the stopping criterion, all \textit{MOC} control parameters were set to their default values selected through iterated F-racing \citep{ref-lopez2016irace} (see Appendix~\ref{default-values}). Notably, we used different datasets for tuning than for the benchmark study.
The stopping criterion was convergence of the HV over 10 generations, with a total maximum of 500 generations.
For all three counterfactual methods, we set the \code{distance\_function} to \code{`gower\_c'} -- a C-based, more efficient version of Gower's distance based on the \pkg{gower} \proglang{R} package \citep{ref-gower2020}.

\begin{figure}[t!]
     \centering
     \includegraphics[width=.8\textwidth]{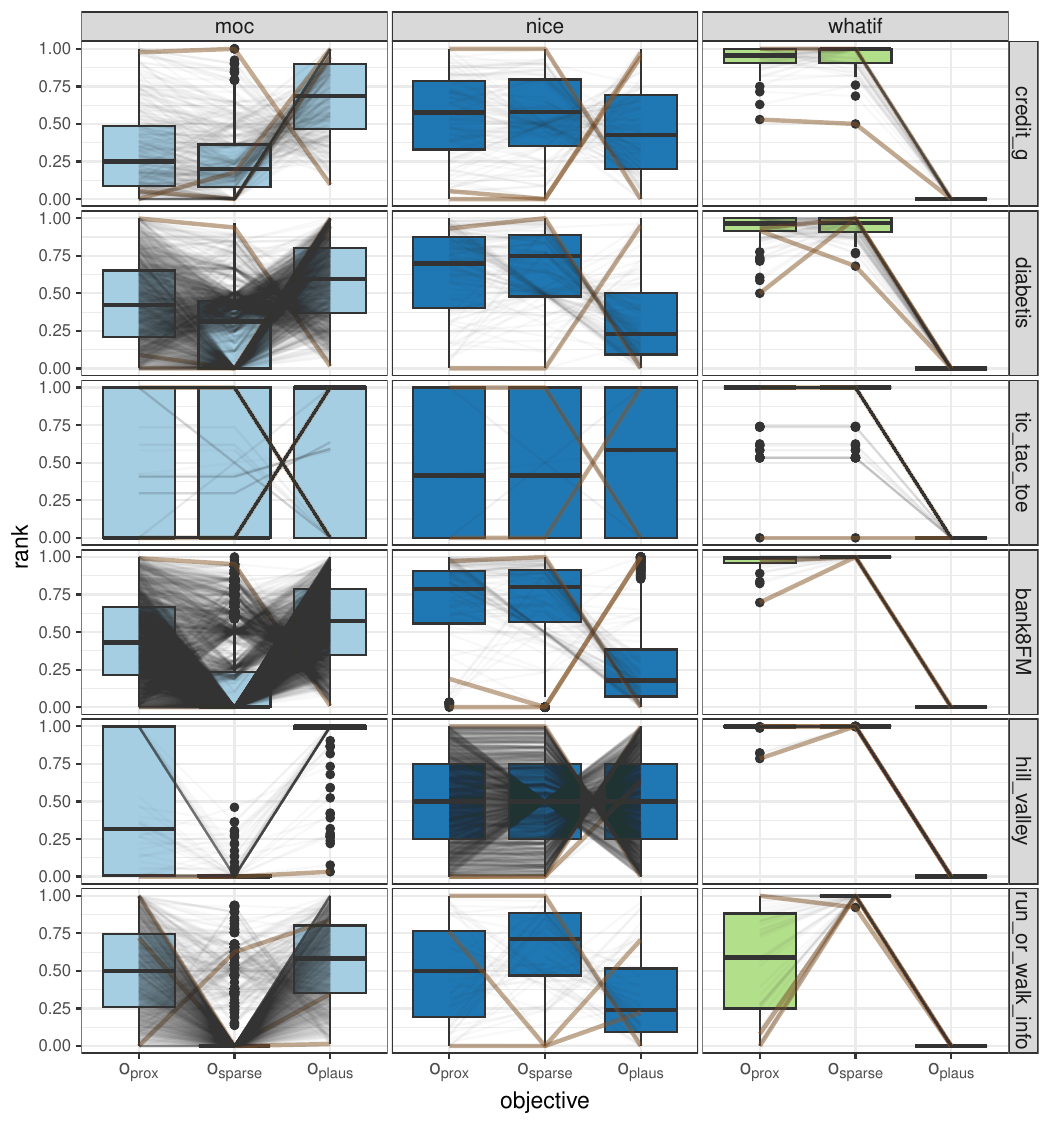}
     \caption{\label{fig:plot-comparison-obj-rank} Comparison of \textit{NICE}, \textit{WhatIf}, and \textit{MOC}  w.r.t.\ their rank in the properties \emph{Proximity} (\ref{proximity}, $o_{\text{prox}}$), \emph{Sparsity} (\ref{sparsity}, $o_{\text{spars}}$) and
\emph{Plausibility} (\ref{plausibility}, $o_{\text{plaus}}$).
Each gray line reflects a counterfactual (for clarity purposes, only a maximum of 2000 counterfactuals are displayed). The counterfactuals with the lowest and therefore best rank in an objective display the brown lines.
     % $^\star$ indicate the significance level of the Wilcoxon rank-sum test that the distribution of ranks for two methods (\textit{MOC} vs. \textit{NICE} and \textit{WhatIf} vs. \textit{NICE}) do not differ ($0 \,\, ^{\star\star\star}  \,\, 0.001 \,\, ^{\star\star} \,\, 0.01 \, ^\star \,\, 0.05 \,\, ns \,\, 1$).
     Lower values are better.
}
\end{figure}

As stated in Section~\ref{sec:methods}, we prefer a set of counterfactuals over a single one. \textit{MOC} is designed to return multiple counterfactuals and we also let \textit{NICE} and \textit{WhatIf} return multiple ones. 
Therefore, the \textit{NICE} control parameter \code{finish\_early} was set to \code{FALSE}, corresponding to our second \textit{NICE} extension (Section~\ref{nice-chapter}). 
In addition, we computed counterfactuals for each of the three different reward functions by varying the \code{optimization} hyperparameter and combined them for a final set of counterfactuals, as recommended in Section~\ref{nice-chapter}.
For \textit{WhatIf}, the number of counterfactual was set to 10  via the \code{n\_counterfactuals} parameter, in accordance with \cite{ref-dandl2020}.
All other \textit{NICE} and \textit{WhatIf} control parameters (except the \code{distance\_function}, see above) were set to their default values (Appendix~\ref{default-values}). 

For the evaluation, we only considered the counterfactuals that (1) achieve the desired prediction such that $o_{\text{valid}} = 0$ and (2) are not dominated by other counterfactuals produced by the same method according to the remaining three objectives ($o_{\text{prox}}$, $o_{\text{sparse}}$ and $o_{\text{plaus}}$).  By design of the three methods, criterion (1) always holds for counterfactuals of \textit{WhatIf} and \textit{NICE} and (2) always for \textit{MOC}.

\begin{figure}[ht]
    \centering
    \includegraphics{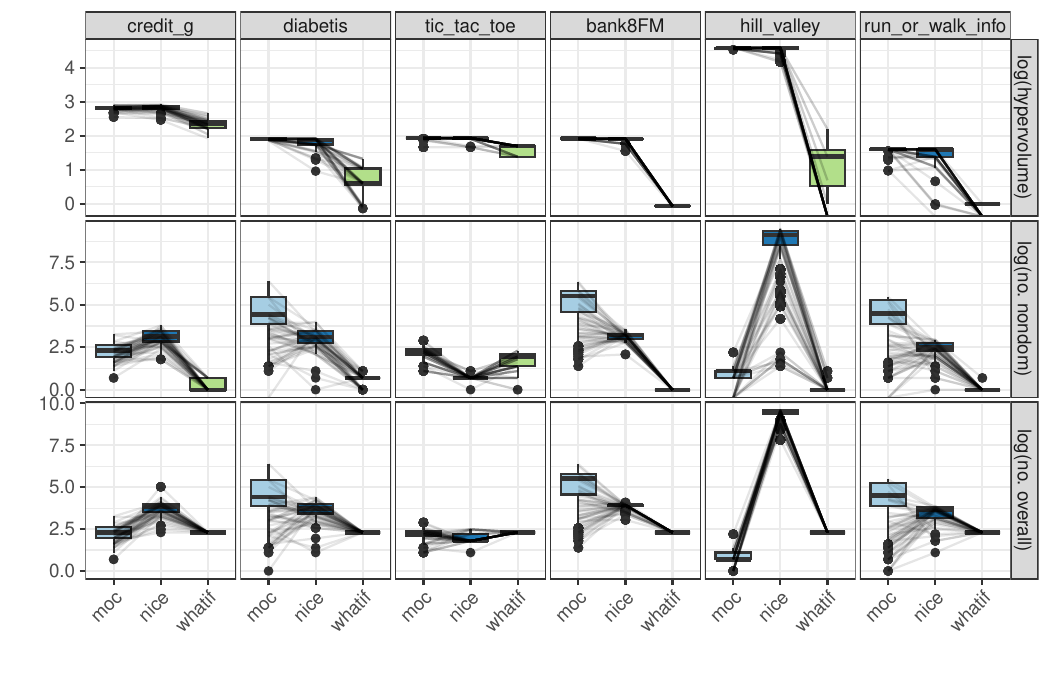}
    \caption{Comparison  of \textit{NICE}, \textit{WhatIf}, and \textit{MOC}  w.r.t.\ their HV, the number of non-dominated and valid counterfactuals (\textit{no.\ nondom}) and the number of all returned counterfactuals (\textit{no.\ overall}). The values were logarithmized. Higher values are better.}
    \label{fig:plot-hv}
\end{figure}

For Research Question~\ref{RQ:1}, we evaluated the generated counterfactuals by means of the desired properties stated in Definition~\ref{def:cfe}: \emph{Validity} (\ref{validity}, $o_{\text{valid}}$), 
\emph{Proximity} (\ref{proximity}, $o_{\text{prox}}$), \emph{Sparsity} (\ref{sparsity}, $o_{\text{spars}}$) and
\emph{Plausibility} (\ref{plausibility}, $o_{\text{plaus}}$).
% Since \textit{MOC} can provide an unlimited amount of counterfactuals, whereas the number for \textit{NICE} and \textit{MOC} is limited by design, we focused on the (maximal) $10$ best counterfactuals for each method for a fair comparison.
%Best means that counterfactuals (1) achieve the desired prediction and (2) are not dominated by other counterfactuals produced by the same method. By design of the three methods, criterion (1) always holds for counterfactuals of \textit{WhatIf} and \textit{NICE} and (2) always for \textit{MOC}.
%If more than ten counterfactuals remained, we used the HV contribution as a tie-breaker for both \textit{MOC} and \textit{NICE}. \textit{WhatIf} never exceeded this number due to the \code{n\_counterfactuals} parameter.
We ranked all counterfactuals per dataset, model, and $\mathbf{x}^{\star}$ by their values in the desired properties, normalized the ranks between $0$ and $1$, and compared the normalized ranks between the methods. 
The ranking ensures that counterfactuals are comparable over all datasets and models.
%We also computed a Wilcoxon rank-sum test, for the null hypothesis that the distributions of the ranks of the counterfactuals for two methods (\textit{MOC} vs. \textit{NICE} and \textit{WhatIf} vs. \textit{NICE}) do not differ.
To take into account all three properties at once, we also computed the HV indicator, which measures the HV in the objective space between the non-dominated counterfactuals and a (worst-case) reference point ($1$ for $o_{\text{prox}}$, no. features for $o_{\text{sparse}}$ and 1 for $o_{\text{plaus}}$).
For Research Question~\ref{RQ:2}, we tracked the runtime behavior for all methods in generating counterfactuals for (row-wise or colum-wise subsets of) the \code{run\_or\_walk\_information} and \code{hill\_valley} datasets.

% As a baseline for \textit{MOC}, we ran random searches with the same population size (20) and number of generations (175) as the \textit{MOC} configurations. The method randomly samples a population of candidates (here, $20\cdot175 = 3500$)  and returns the non-dominated candidates according to the objectives of \textit{MOC} \citep{ref-dandl2020} as counterfactuals.
% We integrated this method also in the \pkg{counterfactuals} package for classification and regression tasks, called \code{RandomSearchClassif} and \code{RandomSearchRegr}, respectively.
%\newpage
\begin{figure}[t!]
     \centering
        \begin{subfigure}[b]{.48\textwidth}
        \centering
            \includegraphics[width = 1\textwidth]{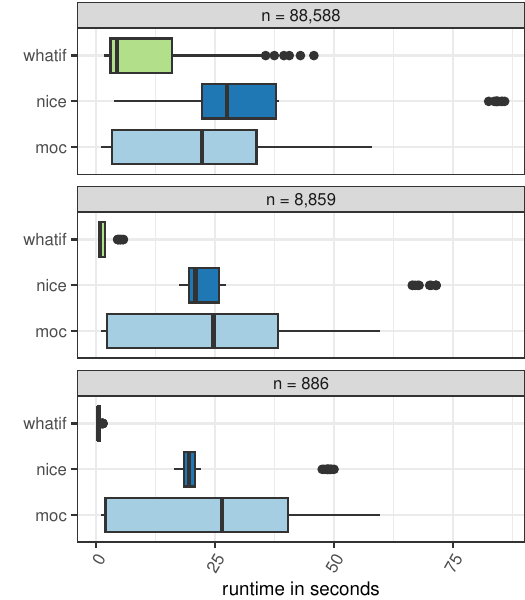} 
            \caption{increasing n (\code{run\_or\_walk\_information})}
        \end{subfigure}
        \hfill
        \begin{subfigure}[b]{.48\textwidth}
        \centering
            \includegraphics[width = 1\textwidth]{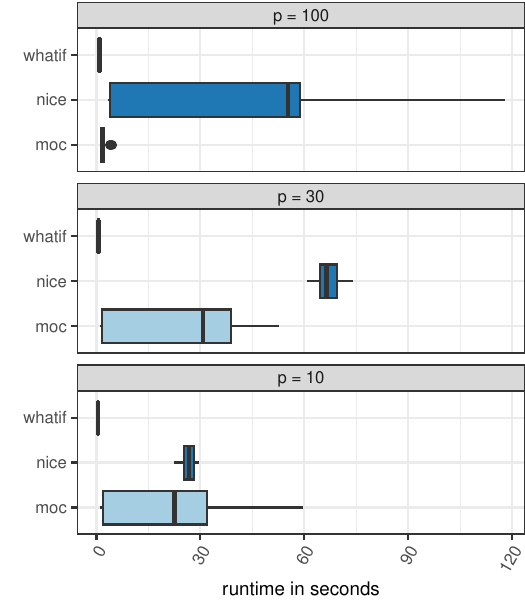} 
            \caption{increasing p (\code{hill\_valley})}
     \end{subfigure}
     \caption{\label{fig:plot-runtime-04}
    Speed comparison of \textit{NICE}, \textit{WhatIf}, and  \textit{MOC} based on row-wise subsets of the \code{run\_or\_walk\_information} dataset and column-wise subsets of the \code{hill\_valley} dataset. The runtimes of \textit{NICE} were aggregated for its three reward function configurations.}
\end{figure}

\subsection{Results}\label{results}

In the following, we present the results for the two stated research questions.

\subsubsection{Research Question 1}
Figure~\ref{fig:plot-comparison-obj-rank} compares the ranking of counterfactuals according to the desired properties for \textit{MOC}, \textit{NICE} and \textit{WhatIf} for each dataset separately.
Figure~\ref{fig:plot-comparison-obj-rank-model} in the Appendix does the same for each model separately. 
Since our setup ensured that all compared counterfactuals achieved the desired prediction, we omitted the results for the first property \emph{Validity} (\ref{validity}, $o_{\text{valid}}$).
 Each gray line reflects a counterfactual. The counterfactuals with the lowest and therefore best rank in one of the three remaining objectives display the brown lines.
Appendix~\ref{benchmarking-results-section-app} shows the results on the property instead of the raking scale for each model and dataset separately. They agree with the results shown here.

WhatIf's counterfactuals changed on average more features ($o_{\text{spars}}$) and had the highest distances to
$\mathbf{x}^{\star}$ ($o_{\text{prox}}$), making \textit{WhatIf} inferior to the other methods w.r.t.\ the desired
counterfactual properties \emph{Sparsity} (\ref{sparsity}) and \emph{Proximity} (\ref{proximity}). However, its counterfactuals have low training data distances ($o_{\text{plaus}}$) by design, guaranteeing \emph{Plausibility} (\ref{plausibility}).  

Compared with \textit{MOC}, the counterfactuals of \textit{NICE} on average changed more features and had often a higher distance to $\mathbf{x}^{\star}$, indicating that \textit{NICE} was overall inferior to \textit{MOC} w.r.t.\ \emph{Sparsity} and \emph{Proximity}. 
However, on average, the counterfactuals of \textit{NICE} had lower training data distances (measuring \emph{Plausibility}) than MOC's counterfactuals.

Figure~\ref{fig:plot-hv}, displays the  HV, the number of non-dominated, valid counterfactuals, and the overall number of returned counterfactuals (including dominated and/or non-valid ones) on the log scale for each dataset and method. 
Overall, \textit{MOC}'s counterfactuals achieved the highest HV closely followed by \textit{NICE}, indicating that \textit{MOC} is slightly superior when considering all objectives simultaneously. 
The HV of \textit{WhatIf}'s counterfactuals is comparably low except for the \texttt{tic\_tac\_toe} dataset with a low number of categorical features.
While all counterfactuals of \textit{MOC} are (by design) non-dominated by other counterfactuals returned by the method, many of the counterfactuals of \textit{NICE} or \textit{WhatIf} are dominated by others generated by the same method. Apart from the \texttt{tic\_tac\_toe} dataset, \textit{WhatIf} produced the least non-dominated counterfactuals. \textit{MOC} generated the most non-dominated counterfactuals except for the \texttt{credit\_g} and \texttt{hill\_valley} datasets.

\subsubsection{Research Question 2}

Figure~\ref{fig:plot-runtime-04} compares the runtimes of our extended \textit{WhatIf} and \textit{NICE} versions with \textit{MOC}. \textit{WhatIf} was the fastest and best scaling method. \textit{NICE} ran on average 17 times longer than \textit{MOC} for high $p$ and almost 1.6 times longer for high $n$. This is because for the \code{hill\_valley} dataset with $p = 100$ features, the method at worse needs to evaluate $(p^2 + p)/2 = 5050$ observations for each of the three reward functions. For low $p$ the differences diminished between \textit{NICE} and \textit{MOC}. For low $n$, \textit{NICE} was on average even faster than \textit{MOC}.

\subsection{Discussion}
In the following, we briefly discuss the suitability of each method for different scenarios based on the
results of our benchmark study.
\textit{MOC} returned on average the most non-dominated counterfactuals of highest-quality when considering all desired properties simultaneously. 
Our extended \textit{NICE} version had comparatively high runtimes for a medium to high number of features.
\textit{WhatIf} was the fastest method, but (by design) its counterfactuals suggested changes to
many features, impeding the interpretation. The method is suitable in time-critical scenarios for datasets with a few categorical features.
%% -- Conclusion ------------------------------------------------------------

\section{Conclusion}\label{conclusion}

In this work, we introduced the \pkg{counterfactuals} \proglang{R} package, which to the best of our knowledge 
is the first \proglang{R} package that provides several counterfactual methods via a unified interface. The package includes the method \textit{MOC} as well as
extended versions of \textit{WhatIf} and \textit{NICE}, which are all capable of returning multiple 
counterfactuals for regression and (binary and multiclass) classification models.
In addition, we illustrated that the \pkg{counterfactuals} package is quickly extensible with new
methods. This is crucial, as the variety of counterfactual methods proposed in research is growing 
rapidly, but the number of implemented methods in \proglang{R} is very limited. Furthermore, the package offers a variety of functionalities for evaluating and visualizing 
the counterfactuals.
Thus, our package facilitates the application of counterfactual methods in practice for auditing machine learning models.

The results of our benchmark study and other research \citep[e.g.,][]{ref-verma2020counterfactual}
suggest that no existing counterfactual method is superior in all situations. This underlines the benefit 
of the \pkg{counterfactuals} package, which makes a variety of methods readily available to the user. 
%Due to the limited number of datasets and evaluation metrics, our benchmark
%results should not be considered general, but rather hypotheses for future benchmark studies. 
Furthermore, the object-oriented concept of our package and the openly available benchmark code allows new methods to easily compete with those currently available. 

%% -- Optional special unnumbered sections -------------------------------------

\section*{Computational details}

The results in this work were obtained using
\proglang{R}~4.2.2 \cite{ref-RCore2021}. 
\proglang{R} itself
and most of the packages used are available from CRAN -- including the \pkg{counterfactuals} \proglang{R} package \citep{ref-counterfactuals2022}. We included all data examples of Sections~\ref{sec:use-cases}~and~\ref{extension-of-the-package} in dedicated vignettes. To facilitate full reproducibility of the benchmark study of Section~\ref{benchmarking}, we created a dedicated Github repository: \url{https://github.com/slds-lmu/benchmark_2022_counterfactuals}. The experiments were run in parallel with the help of the \pkg{batchtools} package \citep{ref-batchtools2017} on a computer with a 2.60 GHz Intel(R) Xeon(R) processor, and 32 CPUs. Training (incl. tuning) the models took 53 hours spread over 15 CPUs, generating the counterfactuals took 37 hours spread over 14 CPUs.

\section*{Acknowledgments}
This work has been partially supported by the Federal Statistical Office of Germany.

%% -- Bibliography -------------------------------------------------------------
%% - References need to be provided in a .bib BibTeX database.
%% - All references should be made with \cite, \citet, \citep, \citealp etc.
%%   (and never hard-coded). See the FAQ for details.
%% - JSS-specific markup (\proglang, \pkg, \code) should be used in the .bib.
%% - Titles in the .bib should be in title case.
%% - DOIs should be included where available.

\bibliography{refs}

%% -- Appendix (if any) --------------------------------------------------------
%% - After the bibliography with page break.
%% - With proper section titles and _not_ just "Appendix".

\newpage

\begin{appendix}

\section{Algorithmic reference}\label{app:algo-ref}

\begin{algorithm}[ht!]
 		\caption{\textit{MOC} based on \cite{ref-dandl2020} as implemented in the \pkg{counterfactuals} \proglang{R} package (Section~\ref{MOC22})}

 		\label{algo:moc}
 		\textbf{Inputs:} \\
 		Data point to explain prediction for $\mathbf{x}^\star \in \mathcal{X}$ \\
		Desired outcome (range) $Y' \subset \mathbb{R}$ \\
 		Prediction function $\hat{f}:\mathcal{X}\rightarrow \mathbb{R}$ \\
 		Observed data $\mathbf{X}$ \\
 		Number of generations $n_{\text{generations}}$ \\
 		Size of population $\mu$ \\
 		Recombination and mutation methods including probabilities \\
 		Selection method and initialization method\\
        Stopping criterion \\
 		(Additional user inputs, e.g., range of numerical features, immutable features, distance function) \\
 		\begin{algorithmic}
 			\item[1:]Initialize population $P_0$ with $|P_0| = \mu$
 			\item[2:] Evaluate candidates according to the four objectives of Equation~\ref{eq:moc-min-prob}
 			\item[3:] Set $t = 0$
 			\item[4:] \textbf{while} stopping criterion not met
 			\item[5:] \hspace{1cm} $C_{t} =$ \texttt{create\_offspring}($P_{t}$), $|C_t| = \mu$ 
 			by selecting, recombinating and mutating \\
 			\hspace{1.1cm} parents with given probabilities
 			\item[6:] \hspace{1cm} Combine parents and offspring $R_t = C_t \cup P_t$
 			\item[7:] \hspace{1cm} Assign candidates to a front according to their objective values: \\
 			\hspace{1.1cm} $(F_1, F_2, ..., F_m) = $ \texttt{ nondominated\_sorting($R_t$)}
 			\item[8:] \hspace{1cm} \textbf{for} $i = 1, ..., m$
 			\item[9:] \hspace{2cm} Sort candidates within a front with (tailored) crowding distance sorting: \\
 			 \hspace{2.3cm} $\tilde{F}_i$ = \texttt{crowding\_distance\_sort($F_i$)}
            \item[10:] \hspace{.9cm} \textbf{end for}
 			\item[11:] \hspace{.9cm} Set $P_{t+1} = \emptyset$ and $i = 1$
 			\item[12:] \hspace{.9cm} \textbf{while} $|P_{t+1}| + |\tilde{F}_i| \le \mu$ 
 			\item[13:] \hspace{2cm} $P_{t+1} = P_{t+1} \cup \tilde{F}_i$
 			\item[14:] \hspace{2cm} i = i + 1
			\item[15:] \hspace{.9cm} \textbf{end while}
			\item[16:] \hspace{.9cm} Choose first $\mu-|P_{t+1}|$ elements of $\tilde{F}_i$: $P_{t+1} = P_{t+1} \cup \tilde{F}_i[1:(\mu-|P_{t+1}|)]$
 			\item[17:] \hspace{.9cm} $t = t + 1$
 		    \item[18:] \textbf{end while}
 		    \item[19:] Return unique, non-dominated candidates of $\bigcup_{k = 0}^{t} P_{k} \setminus \mathbf{x}^\star$
	    \end{algorithmic}
\end{algorithm}  

\begin{algorithm}[ht!]
 		\caption{\textit{NICE} based on \cite{ref-brughmans2021nice} as implemented in the \pkg{counterfactuals} \proglang{R} package}

 		\label{algo:nice}
 		\textbf{Inputs:} \\
 		Data point to explain prediction for $\mathbf{x}^\star \in \mathcal{X}$ \\
		Desired outcome (range) $Y' \subset \mathbb{R}$ \\
 		Prediction function $\hat{f}:\mathcal{X}\rightarrow \mathbb{R}$ \\
 		Observed data $\mathbf{X}$ \\
 		Reward function $R_O$, $O \in $\{sparsity, proximity, plausibility\} \\
 		Indicator whether multiple counterfactuals should be returned $return\_multi$ \\
 		Indicator whether to terminate as soon as desired prediction is reached $finish\_early$ \\
 		(Additional user inputs, e.g., distance function) \\
 		\begin{algorithmic}
 			\item[1:] Find closest observed datapoint $x^{nn} \in \mathbf{X}$ to $\mathbf{x}^\star$ with desired prediction (Equation~\ref{eq:nice_xnn})
 			\item[2:] Set $\mathbf{x}^{best} = \mathbf{x}^\star$
 			\item[3:] Initialize archive set $A = \emptyset$
 			\item[4:] Set $J = \{j \in \{1, ..., p\}: x^{nn}_j \neq x_j^{best}\}$
 			\item[5:] \textbf{while} ($\hat{f}(\mathbf{x}^{best}) \not \in Y'$ \& $finish\_early$ == TRUE) | ($J \neq \emptyset$)
 			\item[6:] \hspace{1 cm} $j^{best} = \emptyset$
 			\item[7:] \hspace{1cm} \textbf{for} $j \in J$:
 			\item[8:] \hspace{2cm} $\mathbf{x} = \mathbf{x}^{best}$
 			\item[9:] \hspace{2cm} Create new candidate by replacing one feature: $x_j = x^{nn}_j$
 			\item[10:] \hspace{1.8cm} \textbf{if} $R_O(\mathbf{x}) > R_O(\mathbf{x}^{best})$: $\mathbf{x}^{best} = \mathbf{x}$ and $j^{best} = j$
 			\item[11:] \hspace{1.8cm} Save created candidate in an archive: $A = A \cup \mathbf{x}$
            \item[12:] \hspace{1cm} \textbf{end for}
 			\item[13:] \hspace{1cm} Update $J = J \backslash\, j^{best}$
 			\item[14:] \textbf{end while}
 			\item[15:] \textbf{if} $return\_multi$: return $\{\mathbf{a} \in A: \hat{f}(\mathbf{a}) \in Y'\}$
 			\item[16:] \textbf{else} return $\mathbf{x}^{best}$
	    \end{algorithmic}
\end{algorithm}  

\newpage

\clearpage

%\section[The counterfactuals R package]{The \pkg{counterfactuals} \proglang{R} package}\label{sec:cf-package-1}
\section{The counterfactuals R package}\label{sec:cf-package-1}

\subsection{Class diagram}\label{class-diagram}

\begin{figure}[ht]
\includegraphics[width=1\linewidth]{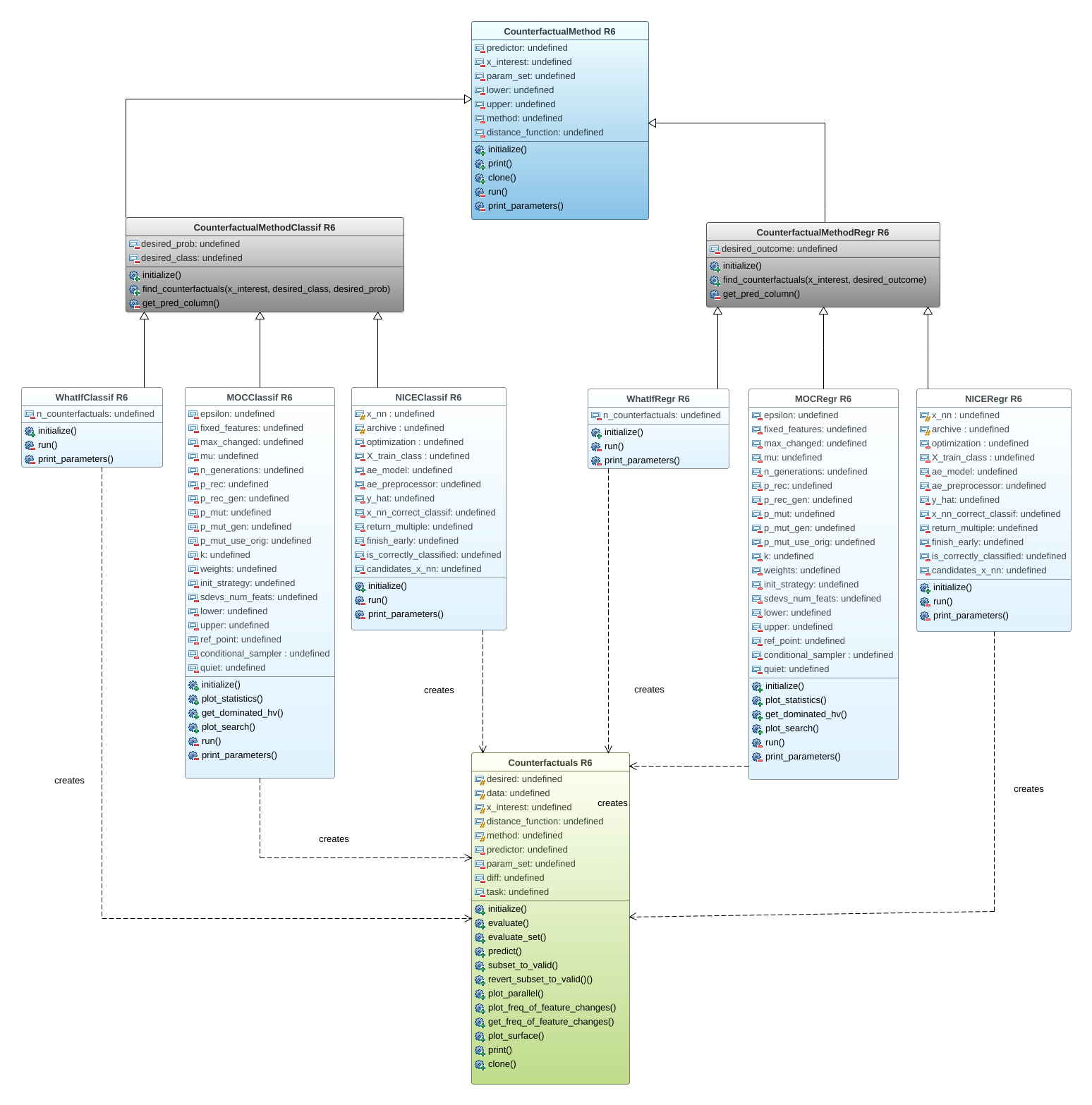} \caption{Detailed class diagram of the \pkg{counterfactuals} package.}\label{fig:class-diagram-detailed}
\end{figure}

\newpage

\subsection{Default values}\label{default-values}
The default parameter settings of the implementations of \textit{WhatIf} and \textit{NICE} should mimic the originally proposed methods in the corresponding papers \citep{ref-wexler2019if,ref-brughmans2021nice}. 
Our \textit{MOC} implementation has the same parameters as the original \textit{MOC} implementation proposed in \citep{ref-dandl2020github} except for \code{p\_rec\_use\_orig}. Instead of resetting after recombination \textit{and} after mutation, we simplify things and reset only once after mutation with a probability of \code{p\_mut\_use\_orig}.
Due to the change in the dependency packages (\code{paradox} and \code{miesmuschel}, see Section~\ref{MOC22}), we re-tuned the \textit{MOC} hyperparameters using the iterated F-race described in \cite{ref-dandl2020} (see Appendix B). The code for tuning can be found here: \url{https://github.com/dandls/moc/tree/irace_newversion}. 
Although tuning identified the usage of the conditional mutator as a successor, we set \code{use\_conditional\_mutator} to \code{FALSE}, since it increases the runtime considerably. 

\begin{table}[ht]
\centering
\begin{tabular}{p{5cm}p{7cm}p{2cm}}
\hline
Name & Description & Default\\ \hline
n\_counterfactuals & The number of counterfactuals to be found. & 1\\ 
lower & Vector of minimum values for numeric features named with the corresponding feature names. If NULL, the element for a numeric feature in lower is taken as its minimum value in observed data. & NULL \\
upper & Vector of maximum values for numeric features named with the corresponding feature names. If NULL, the element for a numeric feature in upper is taken as its maximum value in observed data. & NULL \\ 
distance\_function & Distance function to compute the distances between the original and the training data points. Either the name of an already implemented distance function (`gower' or `gower\_c') or a function. If set to `gower' (default), then Gower's distance (Gower 1971) is used; `gower\_c' is a C-based more efficient version of Gower's distance. A function must have three arguments \code{x}, \code{y}, and \code{data}, and must return a numeric matrix. & `gower' \\\hline
\end{tabular}
\caption{\label{tab:params-whatif}Parameters of \textit{WhatIf} and their default values in the \pkg{counterfactuals} package.}
\end{table}

\newpage

\begin{longtable}{p{5cm}p{7cm}p{2cm}}
\hline
Name & Description & Default\\ \hline
epsilon & If not NULL, candidates whose prediction is farther away from the desired interval than epsilon are penalized. & NULL\\
fixed\_features & Names of features that are not allowed to be changed. NULL (default) allows all features to be changed. & NULL\\
max\_changed & Maximum number of feature changes. NULL (default) allows any number of changes. & NULL\\
mu & The population size. & 20\\
n\_generations & The number of generations. & 175\\
p\_rec & Probability with which an individual is selected for recombination. & 0.71\\
p\_rec\_gen & Probability with which a feature/gene is selected for recombination. & 0.62\\
p\_mut & Probability with which an individual is selected for mutation. & 0.73\\
p\_mut\_gen & Probability with which a feature/gene is selected for mutation. & 0.5\\
p\_mut\_use\_orig & Probability with which a feature/gene is reset to its original value in x\_interest after mutation. & 0.4\\
k & The number of data points to use for the fourth objective (Equation~\eqref{eq:plausibility}). & 1\\
weights & The weights used to compute the weighted sum of dissimilarities for the fourth objective. It is either 
a single value  or a vector of length k summing up to `1` (one weight for each of the $k$ the closest points). NULL (default) means all data points are weighted equally. & NULL\\
lower & Vector of minimum values for numeric features named with the corresponding feature names. If NULL, the element for a numeric feature in lower is taken as its minimum value in observed data. & NULL \\
upper & Vector of maximum values for numeric features named with the corresponding feature names. If NULL, the element for a numeric feature in upper is taken as its maximum value in observed data. & NULL \\ 
init\_strategy & The population initialization strategy. Can be 'random', 'sd', 'traindata' or 'icecurve'. & 'icecurve'\\
use\_conditional\_mutator & Should a conditional mutator be used? The conditional mutator generates plausible
feature values based on the values of the other features. & FALSE\\
distance\_function & 
Distance function for the second and fourth objective. Either the name of an already implemented distance function (`gower' or `gower\_c') or a function. If set to `gower' (default), then Gower's distance (Gower 1971) is used; `gower\_c' is a C-based more efficient version of Gower's distance. A function must have three arguments \code{x}, \code{y}, and \code{data}, and must return a numeric matrix. & `gower' \\ \hline
\caption{Parameters of \textit{MOC} and their default values in the \pkg{counterfactuals} package.}
\label{tab:params-moc} 
\end{longtable}

\begin{longtable}{p{5cm}p{7cm}p{2cm}}
\hline
Name & Description & Default\\ \hline
optimization & The reward function to optimize. Can be 'sparsity' (default), 'proximity', or 'plausibility'. & 'sparsity'\\
x\_nn\_correct & Should only correctly predicted observations be considered for the most similar instance search? & TRUE\\
margin\_correct & \textbf{Only for regression models}. The accepted margin for considering a prediction as "correct". Ignored if x\_nn\_correct = FALSE. If NULL, the accepted margin is set to half the median absolute distance between the true and predicted outcomes in the observed data. & NULL \\
return\_multiple & Should multiple counterfactuals be returned? If TRUE, the algorithm returns all created
instances whose prediction is in the desired interval. & FALSE\\
finish\_early & Should the algorithm terminate after an iteration in which the prediction for the highest reward
instance is in the desired interval. If FALSE, the algorithm continues until x\_nn is recreated. & TRUE\\ 
distance\_function & 
Distance function for computing the distances between the original and the training data points for finding x\_nn. Either the name of an already implemented distance function (`gower' or `gower\_c') or a function. If set to `gower' (default), then Gower's distance (Gower 1971) is used; `gower\_c' is a C-based more efficient version of Gower's distance. A function must have three arguments \code{x}, \code{y}, and \code{data}, and must return a numeric matrix. & `gower' \\ \hline
\caption{\label{tab:params-nice}Parameters of \textit{NICE} and their default values in the \pkg{counterfactuals} package.}
\end{longtable}

\subsection{Different Machine Learning Interfaces}
\label{ap:mlinterfaces}

The \pkg{counterfactuals} \proglang{R} package only allows machine learning models as an input that are instances of an \code{iml::Predictor} object. 
The \code{Predictor} class encapsulates a fitted model together with its underlying (training) data. In Section~\ref{sec:use-cases}, we saw that it works off-the-shelf with models fitted with the \pkg{randomForest} and \pkg{mlr3} \proglang{R} packages \citep{ref-liaw2002randomForest, ref-mlr32019}. 
In this section, we generate counterfactuals for the plasma retinol example of Section~\ref{subsec:regression-task} for models trained with the \pkg{caret}, \pkg{tidymodels} and \pkg{mlr} packages \citep{ref-kuhn2021caret, ref-kuhn2020tidymodels, ref-bischl2016mlr}. 
While all these machine learning interfaces allow training of a variety of models (linear models, model ensembles, etc.), for illustration, we focus on regression trees. Trees are fitted internally with \pkg{rpart} \citep{ref-rpart2019}, such that -- for the sake of completeness -- we also show how to generate counterfactuals for a \pkg{rpart} tree.
For each tree, we generate a counterfactual for the 100th row of the plasma dataset using the \textit{NICE} method. The counterfactual should propose changes such that for the observation a plasma concentration larger than 500 ng/ml is predicted.   

\begin{CodeChunk}
\begin{CodeInput}
R> library("counterfactuals")
R> library("iml")
R> data("plasma", package = "gamlss.data")
R> x_interest = plasma[100L,]
\end{CodeInput}
\end{CodeChunk}

\subsubsection{caret package}

First, we fit a regression tree model with the help of \pkg{caret}. To avoid tuning of the tree, we manually set the only tuning parameter \code{cp} to 0.01 -- the default of the \pkg{rpart} package. 
Then, we initialize an \code{iml::Predictor} object with the fitted model as an input.

\begin{CodeChunk}
\begin{CodeInput}
R> library("caret")
R> treecaret = caret::train(retplasma ~ ., data = plasma[-100L,],
+   method = "rpart", tuneGrid = data.frame(cp = 0.01))
R> predcaret = Predictor$new(model = treecaret, data = plasma[-100L,],
+   y = "retplasma")
R> predcaret$predict(x_interest)
\end{CodeInput}
\end{CodeChunk}

\begin{verbatim}
##   .prediction
## 1         342.92
\end{verbatim}

For the 100th row of the plasma dataset (our \code{x\_interest} or $\mathbf{x}^\star$), we predict a median value of 342.92 -- the same as in Section~\ref{subsec:regression-task}.
Next, we generate counterfactuals by initializing a \code{NICERegr} object with the instantiated \code{Predictor}. 

\begin{CodeChunk}
\begin{CodeInput}
R> nicecaret = NICERegr$new(predcaret, optimization = "proximity", 
+   margin_correct = 0.5, return_multiple = FALSE)
R> nicecaret$find_counterfactuals(x_interest, 
+   desired_outcome = c(500, Inf))
\end{CodeInput}
\end{CodeChunk}

\begin{verbatim}
#> 1 Counterfactual(s) 
#>  
#> Desired outcome range: [500, Inf] 
#>  
#> Head: 
#>    age sex smokstat   bmi vituse calories   fat fiber alcohol cholesterol
#> 1:  46   1        3 35.26      3   2667.5 131.6  10.1       0       550.5     
#>    betadiet retdiet betaplasma
#> 1:     1210    1291        218
\end{verbatim}

Since for all the examples shown in this section, we internally fit a \code{rpart} model to the same data, the prediction and the counterfactual for \code{x\_interest} will be the same. We, therefore, omit the outputs for the prediction and counterfactual for the following machine learning interfaces.

\subsubsection{tidymodels package}

Regression trees of the \pkg{tidymodels} package also work off-the-shelf. However, for classification models, the \code{iml::Predictor} requires a prediction wrapper function (\code{predict.function}) such that class probabilities are returned instead of class labels. For details, the corresponding help page should be consulted. 

\begin{CodeChunk}
\begin{CodeInput}
R> library("tidymodels")
R> treetm = decision_tree(mode = "regression", engine = "rpart") %>% 
  fit(retplasma ~ ., data = plasma[-100L,])
R> predtm = Predictor$new(model = treetm, data = plasma[-100L,], 
+   y = "retplasma")
R> predtm$predict(x_interest)
R> nicetm = NICERegr$new(predtm, optimization = "proximity", 
+   margin_correct = 0.5, return_multiple = FALSE)
R> nicetm$find_counterfactuals(x_interest = x_interest, 
+   desired_outcome = c(500, Inf))
\end{CodeInput}
\end{CodeChunk}

\subsubsection{mlr package}

For the \pkg{mlr} package, the workflow to generate counterfactuals is similar to the one for the \pkg{caret} package. We only need \code{mlr::RegrTask} and \code{mlr::regr.rpart} objects.

\begin{CodeChunk}
\begin{CodeInput}
R> library("mlr")
R> task = mlr::makeRegrTask(data = plasma[-100L,], target = "retplasma")
R> mod = mlr::makeLearner("regr.rpart")
R> treemlr = mlr::train(mod, task)
R> predmlr = Predictor$new(model = treemlr, data = plasma[-100L,], 
+   y = "retplasma")
R> predmlr$predict(x_interest)
R> nicemlr = NICERegr$new(predmlr, optimization = "proximity", 
+   margin_correct = 0.5, return_multiple = FALSE)
R> nicemlr$find_counterfactuals(x_interest = x_interest, 
+   desired_outcome = c(500, Inf))
\end{CodeInput}
\end{CodeChunk}

\subsubsection{rpart package}

For sake of completeness, we also show how to generate counterfactuals for a regression model directly fitted with the \pkg{rpart} package. 

\begin{CodeChunk}
\begin{CodeInput}
R> library("rpart")
R> treerpart = rpart(retplasma ~ ., data = plasma[-100L,])
R> predrpart = Predictor$new(model = treerpart, data = plasma[-100L,], 
+   y = "retplasma")
R> predrpart$predict(x_interest)
R> nicerpart = NICERegr$new(predrpart, optimization = "proximity", 
+   margin_correct = 0.5, return_multiple = FALSE)
R> nicerpart$find_counterfactuals(x_interest = x_interest, 
+   desired_outcome = c(500, Inf))
\end{CodeInput}
\end{CodeChunk}

\subsection{Class FeatureTweakerClassif}\label{ap:featuretweak}

\begin{CodeChunk}
\begin{CodeInput}
R> FeatureTweakerClassif = R6Class("FeatureTweakerClassif", 
+   inherit = CounterfactualMethodClassif,
+   
+   public = list(
+     initialize = function(predictor, ktree = NULL, epsiron = 0.1, 
+       resample = FALSE) {
+       # adds predictor to private$predictor field
+       super$initialize(predictor) 
+       private$ktree = ktree
+       private$epsiron = epsiron
+       private$resample = resample
+     }
+   ),
+   
+   private = list(
+     ktree = NULL,
+     epsiron = NULL,
+     resample = NULL,
+     
+     run = function() {
+       # Extract info from private fields
+       predictor = private$predictor
+       y_hat_interest = predictor$predict(private$x_interest)
+       class_x_interest = names(y_hat_interest)[which.max(y_hat_interest)]
+       rf = predictor$model
+       
+       # Search counterfactuals by calling functions in featureTweakR 
+       rules = getRules(rf, ktree = private$ktree, 
+         resample = private$resample)
+       es = set.eSatisfactory(rules, epsiron = private$epsiron)
+       tweaks = featureTweakR::tweak(
+         es, rf, private$x_interest, label.from = class_x_interest, 
+         label.to = private$desired_class, .dopar = FALSE
+       )
+       return(tweaks$suggest)
+     },
+     
+     print_parameters = function() {
+       cat(" - epsiron: ", private$epsiron, "\n")
+       cat(" - ktree: ", private$ktree, "\n")
+       cat(" - resample: ", private$resample)
+     }
+   )
+ )
\end{CodeInput}
\end{CodeChunk}

\section{Benchmarking}\label{benchmarking-1}

\subsection{Hyperparameter tuning}\label{benchmarking-section}

For hyperparameter tuning, we used random search (with 30 evaluations) and 5-fold CV with the
misclassification error as a performance measure. Table~\ref{tab:hyperparams} shows the tuning search space of each
model. Numerical features were standardized and categorical ones were one-hot encoded using the \pkg{mlr3pipelines} package \citep{ref-mlr3pipelines2021} The optimizer for the neural
network was ADAM \citep{ref-kingma2014adam}, and early stopping was imposed after 5 
patience steps.  
All other hyperparameters were set to their default values in the packages of the mlr3 ecosystem
\citep{ref-mlr32019}. 
For the \texttt{hill\_valley} dataset we used the default deep and wide architecture (two layers) inspired by \cite{ref-ericksonetal2020autogluon} as implemented in the \pkg{mlr3keras} package without tuning \citep{ref-mlr3keras2021}.
Table~\ref{tab:accuracies} shows the accuracies of each model using nested resampling 
(with 5-fold CV in the inner and outer loop).

\begin{table}[H]
\centering
\begin{tabular}{lll}
\hline
Model & Hyperparameter & Range\\ \hline
randomForest & ntrees & {}[0, 1000]\\
xgboost & nrounds & {}[0, 1000]\\
svm & cost & {}[0.01, 1]\\
logreg & - & -\\
neuralnet & lr & {}[0.00001, 0.1]\\
 & layer\_size & {}[1, 20]\\ \hline
\end{tabular}
\caption{\label{tab:hyperparams}Tuning search space of each model. Hyperparameters \code{ntrees} and \code{nrounds} were log-transformed.}
\end{table}

\begin{table}[H]
\centering
\begin{tabular}{llllll}
\hline
dataset & logistic\_regression & neural\_network & ranger & svm & xgboost \\ 
  \hline
  credit\_g & 0.72 & 0.71 & 0.71 & 0.73 & 0.70 \\ 
  diabetes & 0.75 & 0.72 & 0.75 & 0.73 & 0.72 \\ 
  tic\_tac\_toe & 0.97 & 0.98 & 0.95 & 0.79 & 0.98 \\ 
  bank8FM & 0.94 & 0.94 & 0.94 & 0.95 & 0.94 \\ 
  hill\_valley & 0.60 & 0.53 & 0.56 & 0.48 & 0.57 \\ 
  run\_or\_walk\_info & 0.72 & 0.91 & 0.99 & 0.96 & 0.99 \\ 
   \hline
\end{tabular}
\caption{\label{tab:accuracies}Classification accuracies of each model on each dataset. The accuracies were computed using nested resampling with 5-fold CV in the inner and outer loop.}
\end{table}

\clearpage
\newpage

\subsection{Additional results}\label{benchmarking-results-section-app}

\begin{figure}[ht]
     \centering
     \includegraphics[width=.8\textwidth]{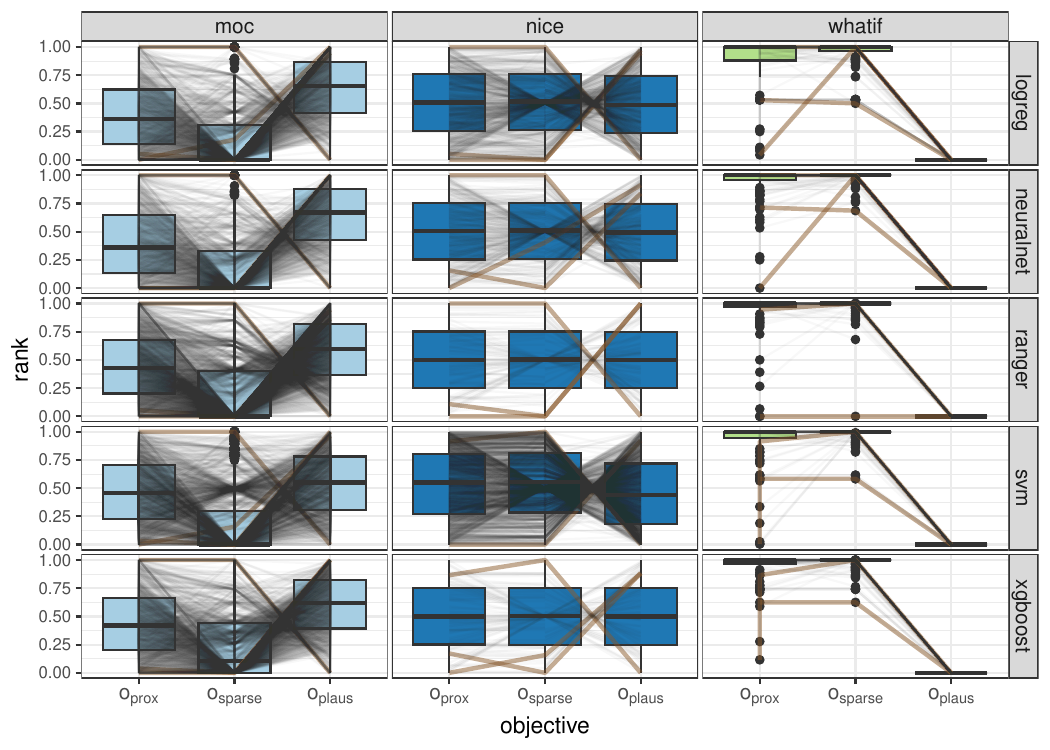}
     \caption{\label{fig:plot-comparison-obj-rank-model} Comparison of \textit{NICE}, \textit{WhatIf}, and \textit{MOC}  w.r.t.\ their rank in the properties \emph{Proximity} (\ref{proximity}, $o_{\text{prox}}$), \emph{Sparsity} (\ref{sparsity}, $o_{\text{spars}}$) and
\emph{Plausibility} (\ref{plausibility}, $o_{\text{plaus}}$). Each gray line reflects a counterfactual (for clarity purposes, only a maximum of 2000 counterfactuals are displayed). The counterfactuals with the lowest and therefore best rank in an objective display the brown lines.
     %$^\star$ indicate the significance level of the Wilcoxon rank-sum test that the distribution of ranks for two methods (\textit{MOC} vs. \textit{NICE} and \textit{WhatIf} vs. \textit{NICE}) do not differ ($0 \,\, ^{\star\star\star}  \,\, 0.001 \,\, ^{\star\star} \,\, 0.01 \, ^\star \,\, 0.05 \,\, ns \,\, 1$). 
     Lower values are better.
}
\end{figure}

\begin{figure}
     \centering
        \begin{subfigure}[b]{1\textwidth}
        \centering
            \includegraphics{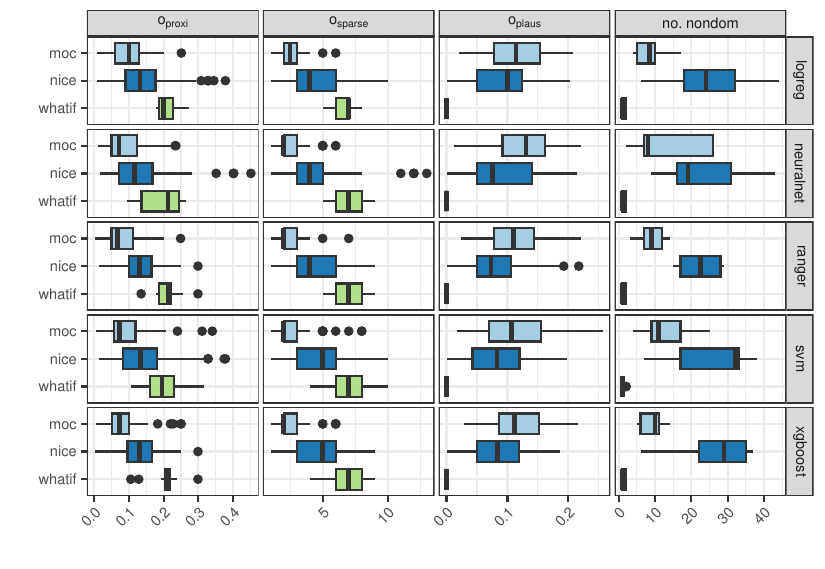} 
            \vspace{-0.5cm}
            \caption{credit\_g}
        \end{subfigure}
        \hfill
        \begin{subfigure}[b]{1\textwidth}
        \centering
             \includegraphics{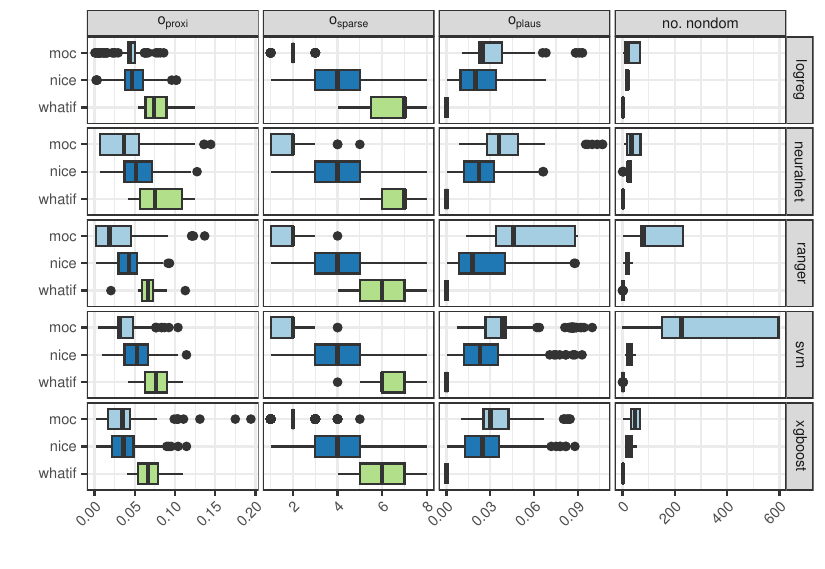} 
            \vspace{-0.5cm}
            \caption{diabetes}
     \end{subfigure}
     \caption{\label{fig:plot-comparison-all}Comparison of \textit{NICE}, \textit{WhatIf}, and \textit{MOC} w.r.t.\ 
    the measures \code{dist\_x\_interest}, \code{no\_changed}, \code{dist\_train} (explained in Section~\ref{sec:use-cases}), and no. nondom (number of non-dominated counterfactuals) for several models for the 
    datasets \code{credit\_g} and \code{diabetes}. $o_{\text{valid}}$ was 0 for all counterfactuals. Lower values are better, except for no. nondom. 
    The figure is based on \cite{ref-dandl2020}.
}
\end{figure}

\begin{figure}
     \centering
     \begin{subfigure}[b]{1\textwidth}
        \centering
            \includegraphics{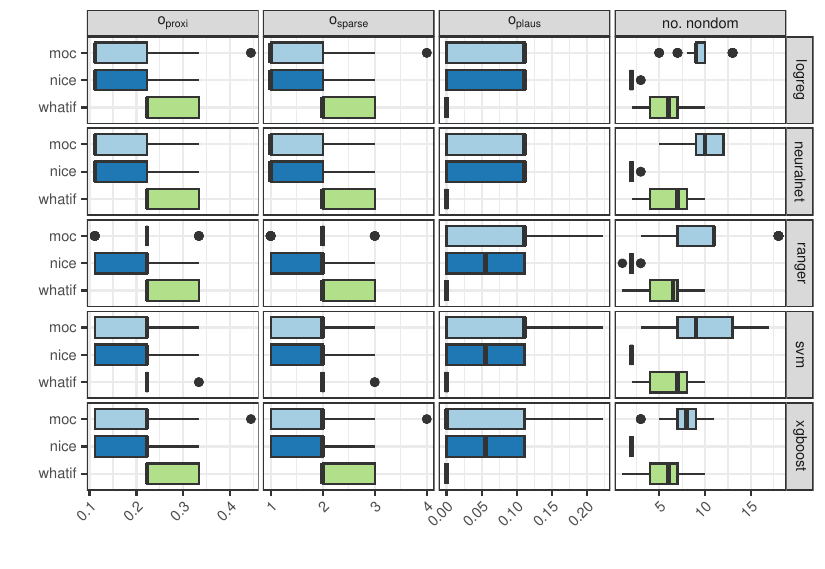} 
            \vspace{-0.5cm}
            \caption{tic\_tac\_toe}
     \end{subfigure}
     \hfill
        \begin{subfigure}[b]{1\textwidth}
        \centering
            \includegraphics{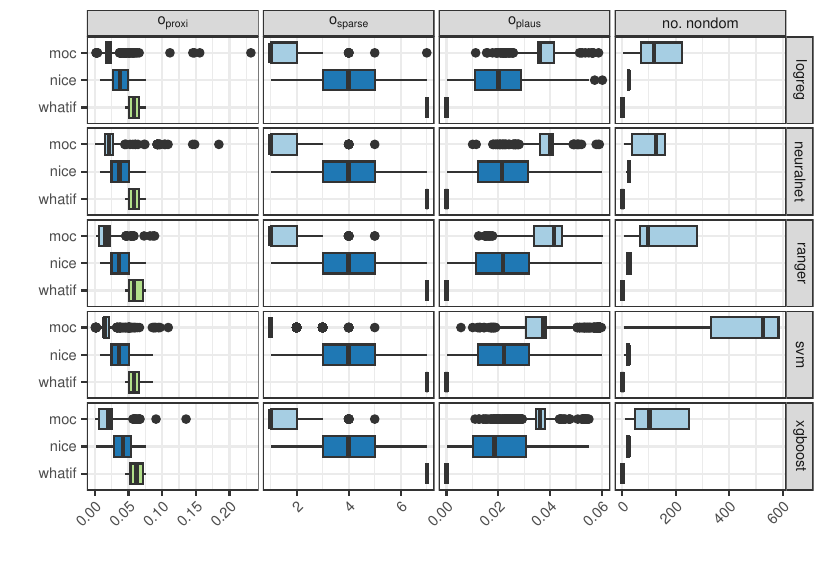} 
            \vspace{-0.5cm}
            \caption{bank8FM}
        \end{subfigure}
\caption{\label{fig:plot-comparison-app-01}Comparison of \textit{NICE}, \textit{WhatIf}, and \textit{MOC} w.r.t.\ 
    the measures \code{dist\_x\_interest}, \code{no\_changed}, \code{dist\_train} (explained in Section~\ref{sec:use-cases}), and no. nondom (number of non-dominated counterfactuals) for several models for the 
    datasets \code{tic\_tac\_toe} and \code{bank8FM}. $o_{\text{valid}}$ was 0 for all counterfactuals. Lower values are better, except for no. nondom. 
    The figure is based on \cite{ref-dandl2020}.}
\end{figure}

\begin{figure}
     \centering
        \begin{subfigure}[b]{1\textwidth}
        \centering
        \begin{subfigure}[b]{1\textwidth}
        \centering
            \includegraphics{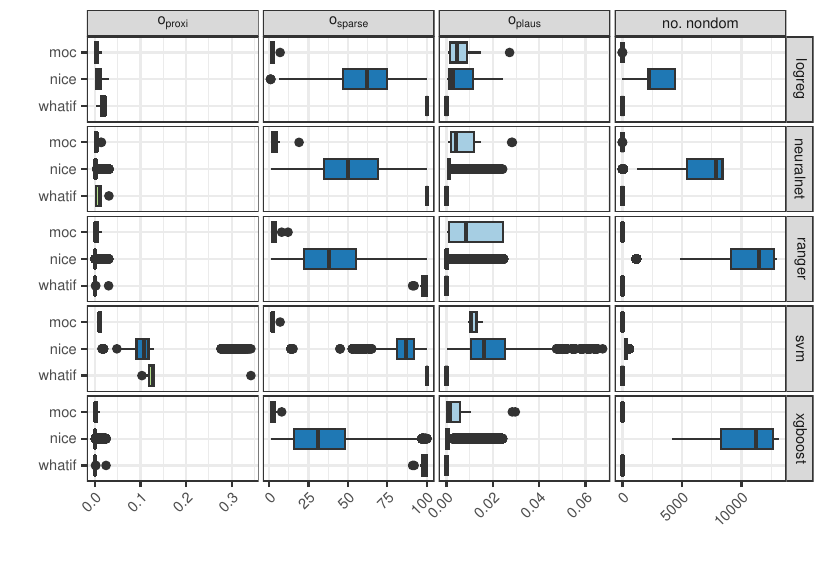} 
            \vspace{-0.5cm}
            \caption{hill\_valley}
                    \hfill
     \end{subfigure}
            \includegraphics{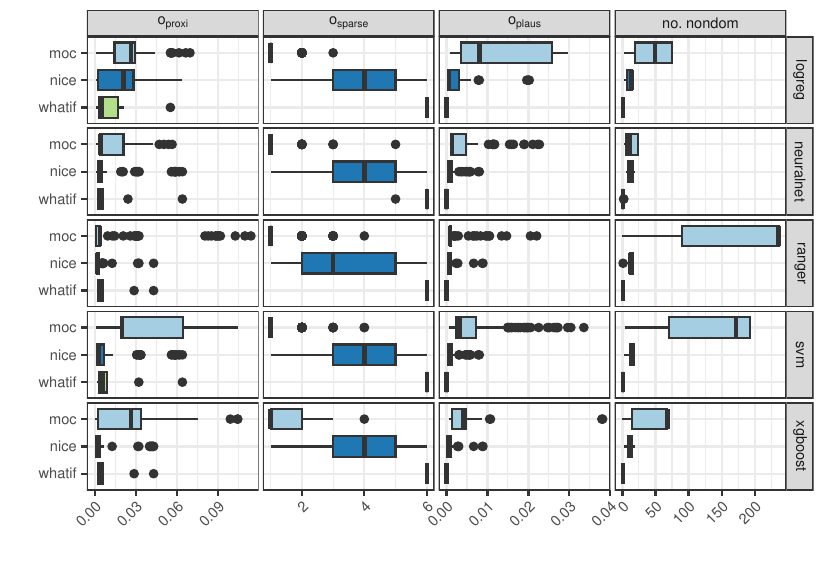} 
            \vspace{-0.5cm}
            \caption{run\_or\_walk\_info}
        \end{subfigure}
        \hfill
        
\caption{\label{fig:plot-comparison-app-02}Comparison of \textit{NICE}, \textit{WhatIf}, and \textit{MOC} w.r.t.\ 
    the measures \code{dist\_x\_interest}, \code{no\_changed}, \code{dist\_train} (explained in Section~\ref{sec:use-cases}), and no. nondom (number of non-dominated counterfactuals) for several models for the 
    datasets \code{hill\_valley} and \code{run\_or\_walk\_information}. $o_{\text{valid}}$ was 0 for all counterfactuals. 
    Lower values are better, except for no. nondom. 
    The figure is based on \cite{ref-dandl2020}.}
\end{figure}

\clearpage
\newpage

\end{appendix}

%% -----------------------------------------------------------------------------

\end{document}